\documentclass[10pt,twocolumn,letterpaper]{article}

\usepackage{iccv}
\usepackage{times}
\usepackage{epsfig}
\usepackage{graphicx}
\usepackage{amsmath}
\usepackage{amssymb}
\usepackage{pifont}%
 \usepackage[table,dvipsnames]{xcolor}
\usepackage{booktabs}
\usepackage{caption}
\usepackage{subcaption}
\usepackage{cuted}
\usepackage{float}
\usepackage[numbers,sort]{natbib}
\usepackage[toc,page,header]{appendix}
\usepackage{minitoc}

\newcommand{\xmark}{\ding{55}}%
\definecolor{aliceblue}{rgb}{0.94, 0.97, 1.0}
\usepackage[accsupp]{axessibility}

\usepackage[pagebackref=true,breaklinks=true,letterpaper=true,colorlinks,bookmarks=false]{hyperref}

\iccvfinalcopy 

\begin{document}

\doparttoc 
\faketableofcontents 



\title{Frozen in Time: A Joint Video and Image Encoder for End-to-End Retrieval}
\author{Max Bain$^{1}$ \quad Arsha Nagrani$^{1}$\footnotemark[2] \quad G\"ul Varol$^{1,2}$ \quad Andrew Zisserman$^{1}$ \\
$^{1}$ Visual Geometry Group, University of Oxford \\
$^{2}$ LIGM, \'Ecole des Ponts, Univ Gustave Eiffel, CNRS \\
{\tt\small \{maxbain, arsha, gul, az\}@robots.ox.ac.uk}
}

\maketitle

\noindent
\vspace{-0.5cm}
\begin{strip}
    \centering\noindent
    \includegraphics[width=0.99\linewidth]{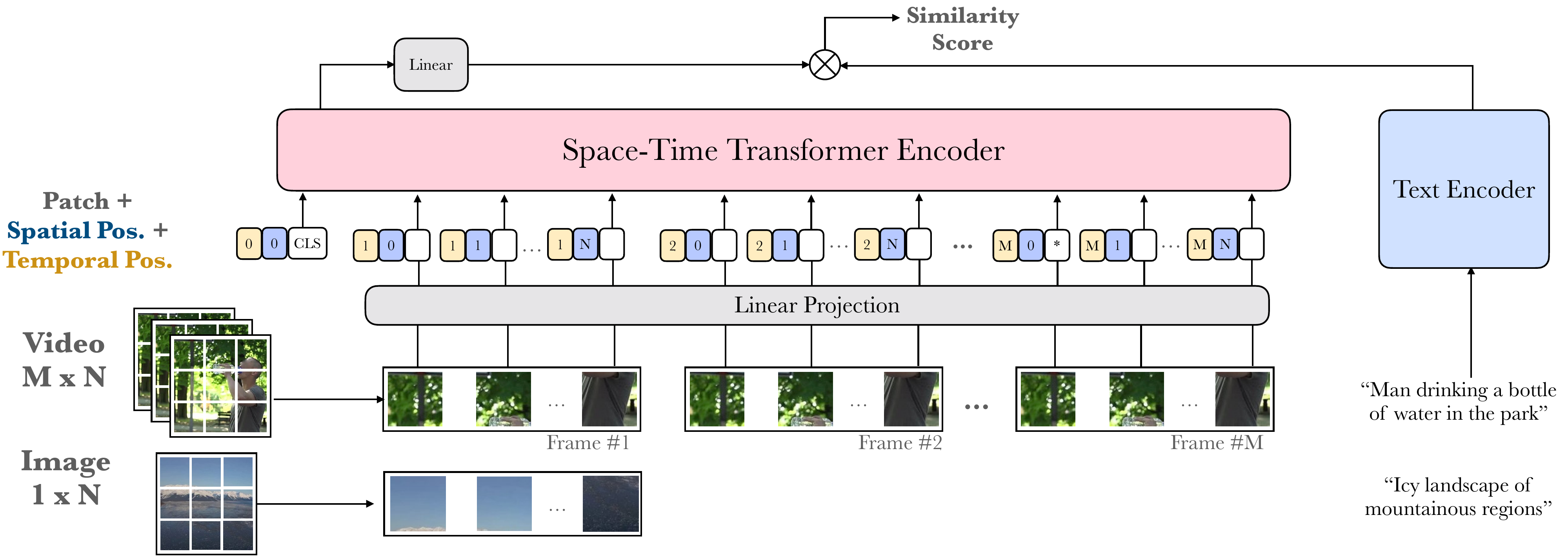}
    \captionof{figure}{\textbf{Joint Image and Video Training:}  Our dual encoding model consists of a visual encoder for images and video and a text encoder for captions. Unlike 2D or 3D CNNs, our space-time transformer encoder allows us to train flexibly on both images and videos with captions jointly, by treating an image as a single frame video.}
    \label{fig:model}
\end{strip}
\renewcommand*{\thefootnote}{\fnsymbol{footnote}}
\footnotetext[2]{Now at Google Research.}
\renewcommand*{\thefootnote}{\arabic{footnote}}
\begin{abstract}
Our objective in this work is video-text retrieval -- in particular a
joint embedding that enables efficient text-to-video retrieval. The
challenges in this area include the design of the visual architecture
and the nature of the training data, in that the available large scale
video-text training datasets, such as HowTo100M, are noisy and hence competitive performance is achieved only at scale through large amounts of compute.

We address both these challenges in this paper.
We propose an end-to-end trainable model that is
designed to take advantage of both large-scale \texttt{image} and
video captioning datasets. Our model is an adaptation and extension of the 
recent  ViT and Timesformer architectures, and consists of attention in both
space and time. The  model is flexible and can be trained on
both image and video text datasets, either independently or in
conjunction.  It  is trained with a curriculum learning schedule
that begins by treating images as `frozen' snapshots of video, and
then gradually learns to attend to increasing temporal context when
trained on video datasets. We also provide a new video-text pretraining dataset WebVid-2M,
comprised of over two million videos with weak captions scraped from
the internet. Despite training on datasets that are an order of magnitude smaller, 
we show that this approach yields state-of-the-art
results on standard downstream video-retrieval benchmarks including MSR-VTT, MSVD, DiDeMo and LSMDC.

\vspace{-2em}
\end{abstract}

\section{Introduction}
Joint visual-text models have become increasingly popular as they enable a 
wide suite of downstream tasks, including text-to-visual retrieval~\cite{lin2014microsoft,wang2016learning,miech18learning,Liu19a},
visual captioning~\cite{vinyals2016show,you2016image,krishna2017dense}, and visual question and answering~\cite{antol2015vqa,lei2018tvqa}. Their rapid development is due to the usual improvements on
three fronts: new neural network architectures (e.g.\ transformers~\cite{vaswani2017attention} for both text and visual inputs); new large-scale
datasets; and new loss functions that are, for example, able to handle label noise~\cite{miech20endtoend}.
However, their development mostly proceeds on two independent tracks: one for {\em images}, with its own architectures,
training datasets and benchmarks~\cite{lin2014microsoft,krishna2017visual,sharma2018conceptual}; and the other for {\em videos} with a similar separation of training datasets and benchmarks~\cite{xu2016msr,anne2017localizing,krishna2017dense,rohrbach2017movie,zhou2018towards,bain2020condensed}. The only common link between the two is that often video networks are initialized by pre-training image networks
on image datasets~\cite{Carreira2017,bertasius2021spacetime}.
This separation of effort is suboptimal given the overlap in information that images and video convey over multiple tasks.
For example, although classifying some human actions requires the temporal ordering of video frames, many actions
can be classified from just their distribution over frames or even from a single frame~\cite{sevilla2021only}.

In this paper we take a step towards unifying these two tracks, by  proposing  a dual encoder architecture which utilises the flexibility of a transformer
visual encoder to train from images-with-captions, from video clips-with-captions, or from both (Fig.~\ref{fig:model}). 
We do this by treating images as a special case of videos that are `frozen
in time'. Using a transformer-based architecture allows us to train
with variable-length sequences, treating
an image as if it was a single frame video, unlike in standard 3D CNNs~\cite{Carreira2017,HaraCVPR2018,xie2018rethinking} where to train on images jointly with videos one must incur the cost
of actually generating a static video.
Furthermore, unlike many recent
methods~\cite{miech18learning,Liu19a, gabeur2020multi} for video-text dual encoding, we do not use a set of `expert networks' that are pre-trained on external image
datasets and then fixed, but instead train the model end-to-end.

This end-to-end training is facilitated by scraping the web for a new large-scale video-text captioning dataset of over two million video alt-text pairs (WebVid-2M). We also take advantage of large-scale image captioning datasets such as Conceptual Captions~\cite{sharma2018conceptual}.  

We make the following contributions: (i) we propose a new 
end-to-end model for video retrieval that does \textit{not} rely on `expert' features, but instead, inspired by~\cite{bertasius2021spacetime} employs a transformer 
architecture with a modified divided space-time attention applied directly to pixels; 
(ii) because our architecture can gracefully handle inputs of different lengths, it is versatile and can be flexibly trained on both video and image datasets (by treating images as a single-frame video). We build on this flexibility by designing a curriculum learning schedule that begins with images and then gradually learns to attend to increasing temporal context when trained on video datasets through temporal embedding interpolation. We show that this increases efficiency, allowing us to train models with far less GPU time; 
(iii) we introduce a new dataset called WebVid-2M, consisting of 2.5M video-text pairs scraped from the web; and finally (iv) we achieve state-of-the-art performance by only using the video modality on MSR-VTT~\cite{xu2016msr}, MSVD~\cite{chen2011collecting},
DiDeMo~\cite{anne2017localizing}
and LSMDC~\cite{rohrbach2017movie} --
outperforming works that use pre-extracted experts from multiple modalities, as well as those that are pretrained on the noisy HowTo100M, which is 20x larger than our dataset in the number of video-text pairs.

\section{Related Works}
\noindent\textbf{Pretraining for video-text retrieval.}
Given that most video-text retrieval datasets tend to be small-scale, the dominant paradigm for video retrieval has been to use a combination of pre-extracted features from `expert' models, including models trained for various diverse tasks and on multiple modalities such as face, scene and object recognition, action classification and sound classification. MoEE~\cite{miech18learning}, CE~\cite{Liu19a}, MMT~\cite{gabeur2020multi} and concurrent work HiT~\cite{liu2021hit} all follow this paradigm, with the overall similarity for a video-text pair obtained as a weighted sum of each expert’s similarity with the text. 

However, since the release of the HowTo100M dataset~\cite{miech2019howto100m}, a large-scale instructional video dataset, there has been a flurry of works leveraging large-scale pretraining to improve video-text representations for tasks such as video question-answering~\cite{seo2021look}, text-video retrieval~\cite{patrick2020support} and video captioning~\cite{zhou2018end}. Although semantically rich and diverse, text supervision from instructional videos is extremely noisy, and hence incurs a large computational cost, as scale is required for competitive results. A few approaches have been proposed to combat the noise --  \eg using loss functions such as MIL-NCE~\cite{miech20endtoend} or using the raw audio~\cite{alayrac2020self,rouditchenko2020avlnet} directly to increase robustness. Given the large size of existing image-captioning datasets, some have naturally tried to overcome the lack of video-caption training data with joint image-text pretraining (such as in MoEE~\cite{miech18learning} and ClipBERT~\cite{lei2021less}). MoEE~\cite{miech18learning} trains on images jointly by feeding in zeros to all expert streams that require videos, such as the motion and audio features, while ClipBERT~\cite{lei2021less} restricts their feature extractors to 2D CNNs. Instead we propose an elegant transformer-based encoder that works well with either images or videos and can be trained effectively on both.

Similar to our work, although only suitable for images is CLIP~\cite{radford2021learning}, which learns an effective joint image-text representation from millions of text-image pairs scraped from the internet using contrastive loss.

\noindent\textbf{End-to-end video representation learning.}
A large number of architectural developments have been driven by action recognition
on datasets such as Kinetics~\cite{Kinetics} where manual labelling has been relatively easier than obtaining textual descriptions for datasets. For a long time this space was dominated by spatio-temporal CNNs such as I3D~\cite{Carreira2017},
3D ResNets~\cite{HaraCVPR2018}, S3D~\cite{xie2018rethinking}
or `R(2+1)D' CNNs \cite{Tran2018ACL}. Here, images are used simply to initialise video models, through inflation~\cite{Carreira2017}. Multigrid scheduling has been proposed for efficient training~\cite{Wu2020AMM}.

\noindent\textbf{Transformers for vision.} A number of works use self-attention for images, either in combination with convolutions~\cite{wang2018non,vaswani2017attention,hu2017relation,carion2020endtoend} or even replacing them entirely. 

Works that use only self-attention blocks tend to apply them at an individual pixel level~\cite{parmar2018image,ramachandran2019stand,cordonnier2019relationship}, often requiring tricks to ensure computational tractability, including restricting the scope of self-attention to a local neighbourhood~\cite{ramachandran2019stand}, adding global self-attention on heavily downsized versions, or sparse key-value sampling~\cite{child2019generating}. To increase efficiency, ViT~\cite{dosovitskiy2021an} decompose images into a sequence of patches and then feeds linear embeddings of
these patches as inputs to a transformer, effectively adding a single convolutional layer to the image at the start. This idea has been extended in DeiT~\cite{touvron2020deit}. For video, previous works also employ self-attention blocks together with CNN layers, for action recognition~\cite{girdhar2017actionvlad} and video classification~\cite{chen20182}. 

In contrast, our architecture consists entirely of self-attention units and is heavily inspired by ViT~\cite{dosovitskiy2021an} and particularly the Timesformer~\cite{bertasius2021spacetime}, which uses divided space and time attention. Unlike these works, we use expandable temporal embeddings to allow flexible training of variable-length videos and images both jointly and separately. We are unaware of any previous works that use self-attention to train on both images and videos in the same model.

\section{Method}

In this section, we describe our transformer-based spatio-temporal model architecture
(Section~\ref{subsec:architecture}), and our training strategy (Section~\ref{subsec:training}).

\subsection{Model Architecture}
\label{subsec:architecture}

\noindent\textbf{Input.} The visual encoder takes as input an image or video clip $X \in \mathrm{R}^{M \times 3 \times H \times W}$ consisting of $M$ frames of resolution $H \times W$, where $M=1$ for images. The text encoder takes as input a tokenised sequence of words.

\noindent\textbf{Spatio-temporal patches.} Following the protocol in ViT and Timesformer~\cite{bertasius2021spacetime}, the input video clip is divided into $M \times N$ non-overlapping spatio-temporal patches of size $P \times P$, where $N = HW/P^2$.

\noindent\textbf{Transformer input.}
The patches $\boldsymbol x \in \mathrm{R}^{M \times N \times 3 \times P \times P}$ are fed through a 2D convolutional layer and the output is flattened, forming a sequence of embeddings $\boldsymbol z\in \mathrm{R}^{MN \times D}$ for input to the transformer, where $D$ depends of the number of kernels in the convolutional layer.

Learned temporal and spatial positional embeddings, $\boldsymbol E^{s} \in \mathrm{R}^{N \times D}$, $\boldsymbol E^{t} \in \mathrm{R}^{M \times D}$ are added to each input token:
\begin{equation}
    {\boldsymbol z^{(0)}_{p,m}} = \boldsymbol z_{p,m} + \boldsymbol E^{s}_{p} + \boldsymbol E^{t}_{m},
\end{equation}
such that all patches within a given frame $m$ (but different spatial locations) are given the same temporal positional embedding $E^{t}_{m}$, and all patches in the same spatial location (but different frames) are given the same spatial positional embedding $ E^{s}_{p}$. Thus enabling the model to ascertain the temporal and spatial position of patches.
 
In addition, a learned [CLS] token~\cite{devlin2019bert} is concatenated to the beginning of the sequence, which is used to produce the final visual embedding output embedding of the transformer.

\noindent\textbf{Space-time self-attention blocks.}
The video sequence is fed into a stack of space-time transformer blocks. We make a minor modification to the Divided Space-Time attention introduced by~\cite{bertasius2021spacetime}, by replacing the residual connection between the block input and the temporal attention output 
with a residual connection between the block input and the spatial attention output (see Section~\ref{sec:spacetime_block} of the Appendix for details). Each block sequentially performs temporal self-attention and then spatial self-attention on the output of previous block. The video clip embedding is obtained from the [CLS] token of the final block.

\noindent\textbf{Text encoding.}
The text encoder architecture is a multi-layer bidirectional transformer encoder, which has shown great success in natural language processing tasks~\cite{devlin2019bert}. For the final text encoding, we use the [CLS] token output of the final layer.

\noindent\textbf{Projection to common text-video space.}
Both text and video encodings are projected to a common dimension via single linear layers. We compute the simliarity between text and video by performing the dot product between the two projected embeddings.

\noindent\textbf{Efficiency.} Our model has independent dual encoder pathways (such as in MIL-NCE~\cite{miech20endtoend} and MMV networks~\cite{alayrac2020self}), requiring only the dot product between the video and text embeddings. This ensures retrieval inference is of trivial cost since it is indexable, i.e.\ it allows application of fast approximate nearest neighbour search, and is scalable to very large scale retrieval at inference time. Given $t$ text queries and $v$ videos in a target gallery, our retrieval complexity is $O(t+v)$. In contrast, ClipBERT~\cite{lei2021less} which inputs both text and video as input to a single encoder, has retrieval complexity $O(tv)$ since every text-video combination must be inputted to the model. Other expert-based retrieval methods such as MoEE~\cite{miech18learning}, CE~\cite{Liu19a} and MMT~\cite{gabeur2020multi} also contain a dual encoder pathway, however they still require query-conditioned weights to compute the similarity scores for each expert, while our model does not.


\subsection{Training Strategy}
\label{subsec:training}

\noindent\textbf{Loss.}
We employ~\cite{Zhai2019ClassificationIA} in a retrieval setting, where matching text-video pairs in the batch are treated as positives, and all other pairwise combinations in the batch are treated as negatives. We minimise the sum of two losses,
video-to-text
and text-to-video:
\begin{equation}
\label{eq:loss1}
    L_{v2t} = -\frac{1}{B}\sum_i^B\log{\frac{\exp(x_i^\top y_i / \sigma)}{\sum_{j=1}^{B} \exp(x_i^\top y_j / \sigma)}}
\end{equation}
\begin{equation}
\label{eq:loss2}
    L_{t2v} = -\frac{1}{B}\sum_i^B\log{\frac{\exp(y_i^\top x_i / \sigma)}{\sum_{j=1}^{B} \exp(y_i^\top x_j / \sigma)}}
\end{equation}
where $x_i$ and $y_j$ are the normalized embeddings of $i$-th video and the $j$-th text respectively in a batch of size $B$ and $\sigma$ is the temperature.

\noindent\textbf{Joint image-video training.} In this work, we train jointly on both image-text pairs as well as video-text pairs, taking advantage of both for larger-scale pretraining. Our joint training strategy involves alternating batches between the image and video datasets. Since the attention mechanism scales with the square of input frames $O(M^2)$, the alternate batch training allows the image batches ($M=1$) to be far greater in size. 

\noindent\textbf{Weight initialisation and pretraining.}
Following~\cite{bertasius2021spacetime}, we initialise the spatial attention weights in the space-time transformer model with ViT~\cite{dosovitskiy2021an} weights trained on ImageNet-21k, and initialise the temporal attention weights to zero.
The residual connections mean that under these initialisation settings, the model is at first equivalent to ViT over each input frame -- thereby allowing the model to learn to attend to time gradually as training progresses. Since transformer architectures have demonstrated most of their success from large-scale pretraining, we utilise two large-scale text-image/video datasets with a joint training strategy, resulting in large improvements in performance.

\noindent\textbf{Temporal curriculum learning.} The space-time transformer architecture allows a variable length input sequence and therefore a variable number of input video frames. If the model has only trained on videos up to length $m$ however, then the temporal positional embedding $\boldsymbol E^t$ will only be learned up to $\boldsymbol E^t_{:m}$.
Therefore, applying the model to input video of sequences up to length $M$ will result the addition of $ \boldsymbol E^t_{m:M}$, which would not yet be learned.

Two temporal expansion methods are investigated: \textit{interpolation} and \textit{zero-padding}. Zeros can be filled in, $ \boldsymbol 0  \rightarrow \boldsymbol E^{t}_{m:M}$, allowing the model to learn the additional temporal positions from scratch during training. Alternatively, interpolation could be used to upsample the temporal embeddings in the temporal dimension, $\boldsymbol E^t_{:m} \rightarrow \boldsymbol E^t_{:M}$. We investigate two methods of interpolation: nearest neighbour and bilinear. The effects of these different initialisations can be found in the Appendix, Section~\ref{sec:temp_expansion}.

We employ this expansion strategy in order to perform curriculum learning in the number of input frames. Initially training on fewer frames has drastic savings in computation, whilst having comparable or even better performance (see Section~\ref{subsec:curriculumexp}).

\noindent\textbf{Frame sampling.} Given a video containing $L$ frames, we subdivide it into $M$ equal segments where $M$ is the desired number of frames for the video encoder. During training, we sample a single frame uniformly from each segment (in a similar manner to  TSN~\cite{wang_tsn} and GST~\cite{gst}).
At test time, we sample the $i^{th}$ frame in every segment, to get a video embedding $v_i$. The values for $i$ are determine using a stride $S$, resulting in an array of video embeddings $\boldsymbol v = [v_0, v_{S}, v_{2S} , v_{M}]$. The mean of these video embeddings is used as the final embedding for the video.

\begin{figure*}
\captionsetup[subfigure]{labelformat=empty}
     \centering
     \begin{subfigure}[t]{0.23\textwidth}
         \centering
         \includegraphics[width=\textwidth]{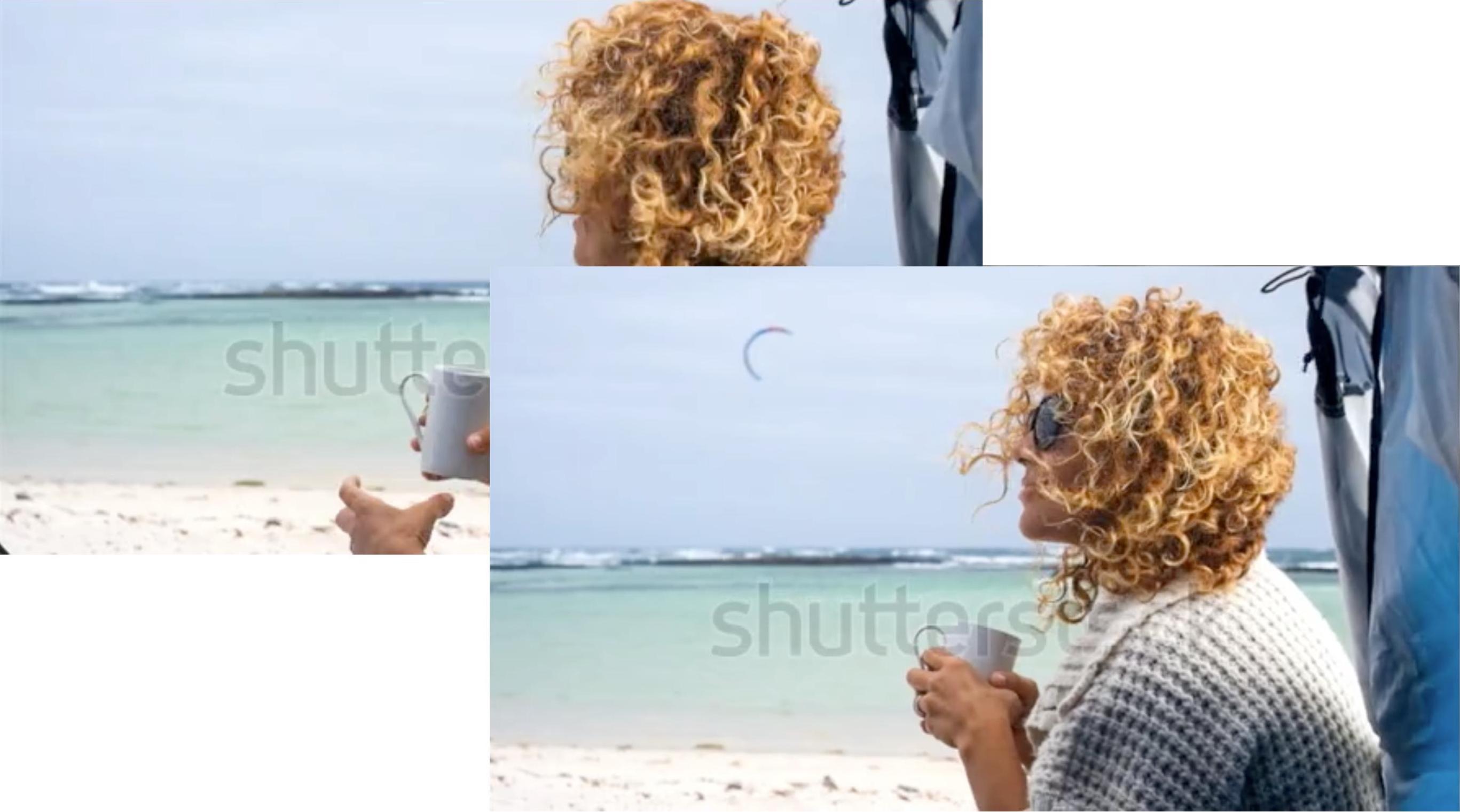}
         \caption{\footnotesize{``Lonely beautiful woman sitting on the tent looking outside. wind on the hair and camping on the beach near the colors of water and shore. freedom and alternative tiny house for traveler lady drinking"}}
     \end{subfigure}
     \hfill
     \begin{subfigure}[t]{0.23\textwidth}
         \centering
\includegraphics[width=\textwidth]{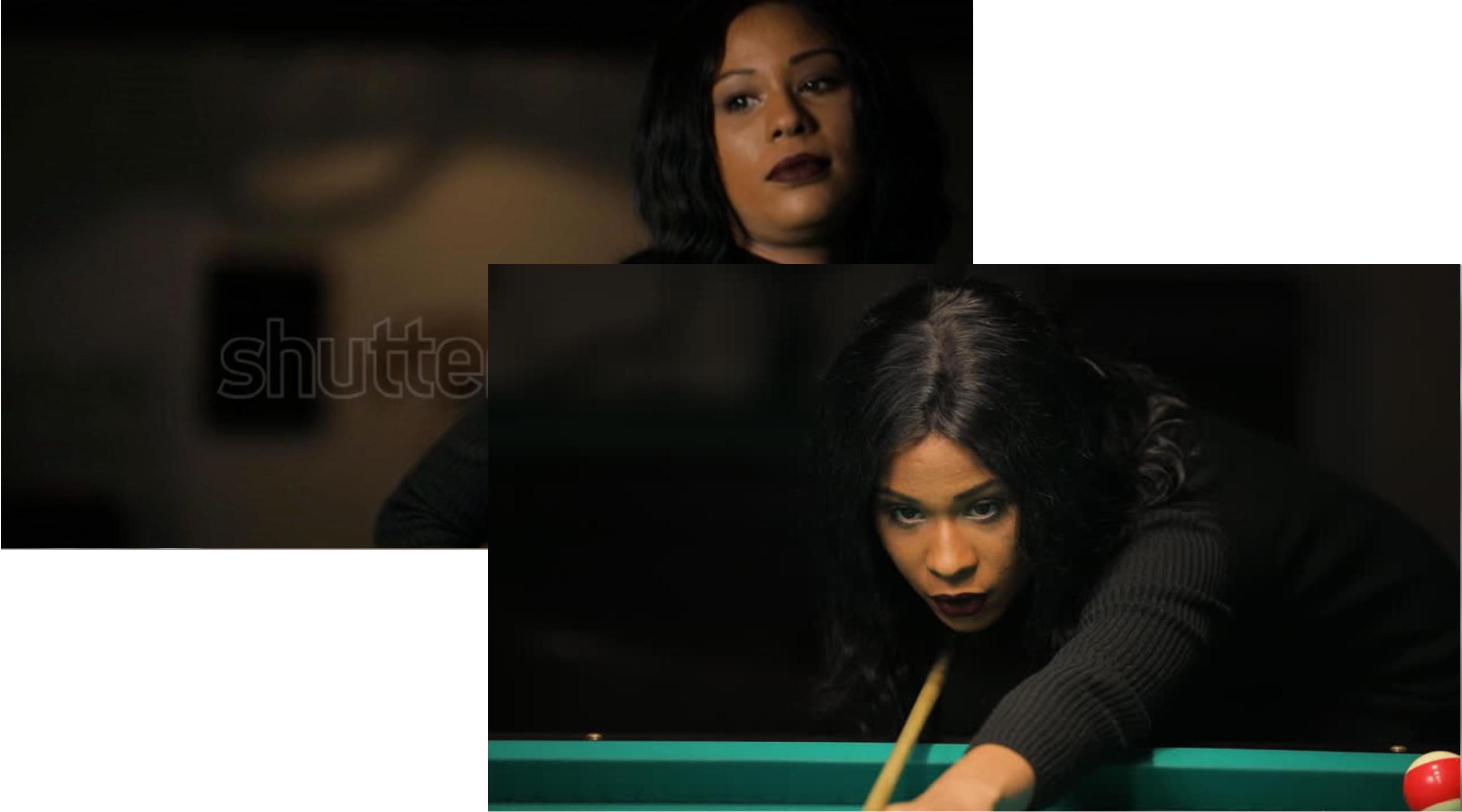}
         \caption{\footnotesize{``Billiards, concentrated young woman playing in club"}}
     \end{subfigure}
     \hfill
     \begin{subfigure}[t]{0.23\textwidth}
         \centering
         \includegraphics[width=\textwidth]{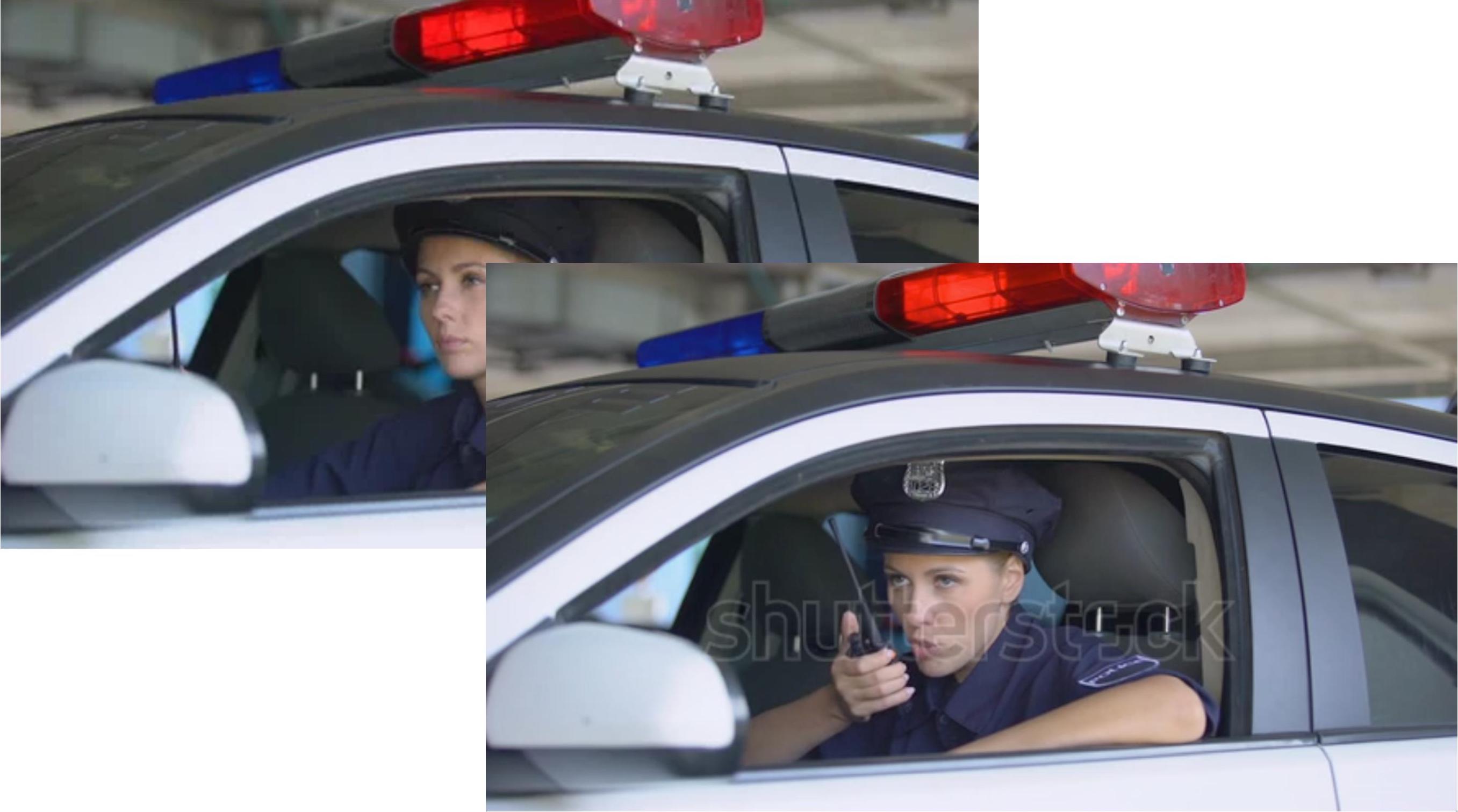}
         \caption{\footnotesize{``Female cop talking on walkie-talkie, responding emergency call, crime prevention''}}
     \end{subfigure}
     \hfill
    \begin{subfigure}[t]{0.23\textwidth}
         \centering
         \includegraphics[width=\textwidth]{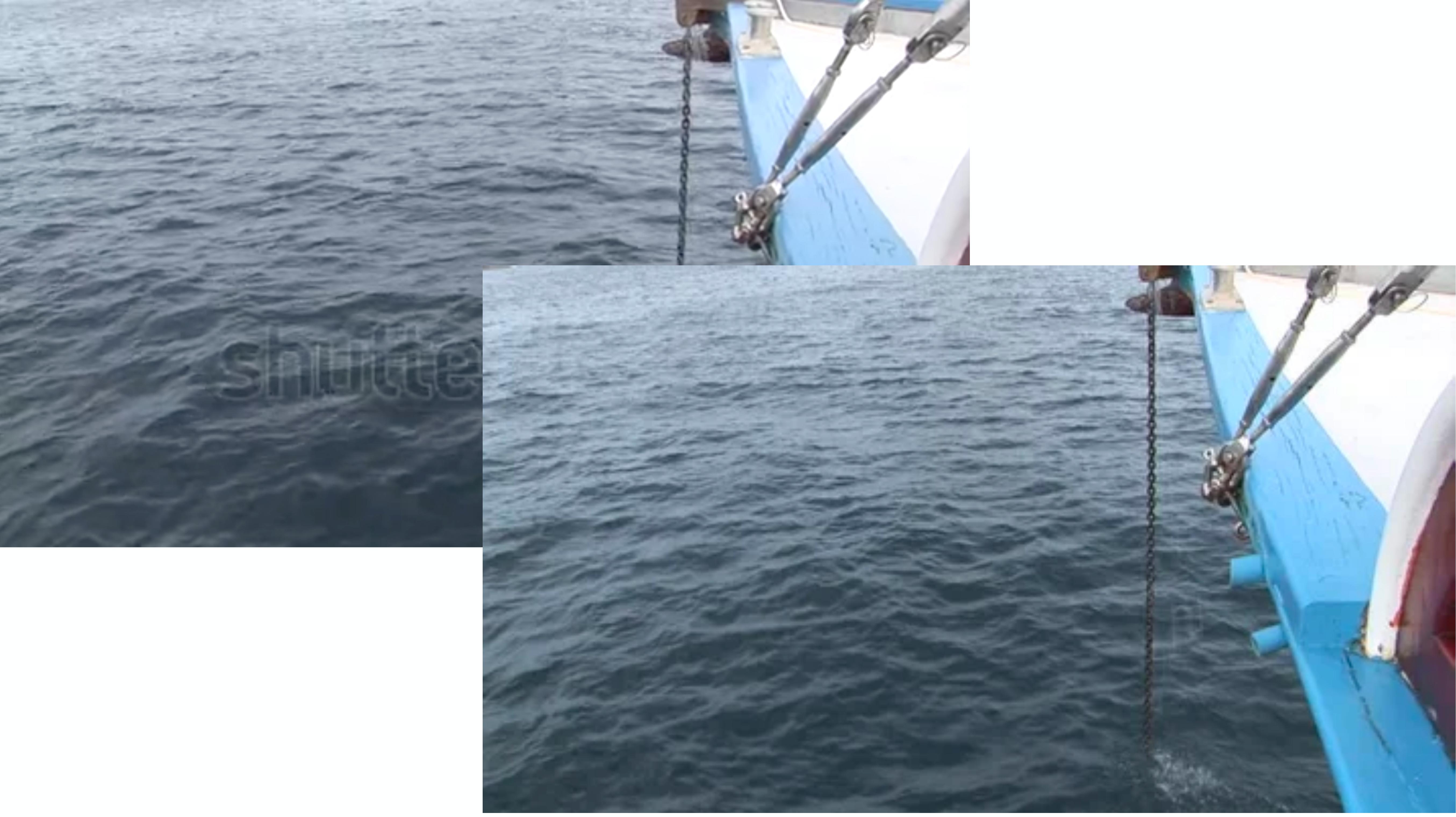}
         \caption{\footnotesize{``Get anchor for departure safari dive boat scuba diving maldives"}}
     \end{subfigure}
     \hfill
                  \caption{\textbf{Example video-caption pairs from the WebVid2M dataset:} Note the different captioning styles: from left to right, captions can be (i) long, slightly poetic, with disjoint sentences and phrases, (ii) succint and to the point, (iii) have a less defined sentence structure with keywords appended to the end, (iv) mention specific places (`maldives'). We show two randomly sampled frames for each video.}

        \label{fig:webvid2m}
\end{figure*}
\section{Experiments}
We first describe the pretraining datasets including our WebVid-2M video-text dataset 
(Section~\ref{subsec:datasets1}), followed by the downstream datasets used for the evaluations
in our experiments (Section~\ref{subsec:datasets2}).
We then describe implementation details of our model (Section~\ref{subsec:implementation}).
Next, we ablate various training components on the MSR-VTT dataset,
in particular the effects of pretraining and our space-time attention modification (Section~\ref{subsec:ablation}),
and our proposed curriculum strategy (Section~\ref{subsec:curriculumexp}).
Then, we compare to the state of the art on
four benchmarks: MSR-VTT, MSVD, DiDeMo and LSMDC (Section~\ref{subsec:sota}).

\subsection{Pretraining Datasets}
\label{subsec:datasets1}
We jointly pretrain our model on image and video data.

\noindent\textbf{Video pretraining: The WebVid-2M Dataset.}
We scrape the web for a new dataset of videos with textual description annotations, called WebVid-2M. Our dataset consists of 2.5M video-text pairs, which is an order of magnitude larger than existing video captioning datasets (see Table~\ref{tab:datastats}).

The data was scraped from the web following a similar procedure to Google Conceptual Captions~\cite{sharma2018conceptual} (CC3M). We note that more than 10\% of CC3M images are in fact thumbnails from videos, which motivates us to use such video sources to scrape a total of 2.5M text-video pairs.
The use of data collected for this study is authorised via the Intellectual Property Office’s Exceptions to Copyright for Non-Commercial Research and Private Study\footnote{\url{www.gov.uk/guidance/exceptions-to-copyright/}}. We are currently performing further analysis of the dataset on its diversity and fairness.

Figure~\ref{fig:webvid2m} provides sample video-caption pairs. There are a variety of different styles used in caption 
creation, as can be seen from Figure~\ref{fig:webvid2m} (left to right) where the first video has a longer, poetic 
description compared to the succinct description for the second video. The third video caption has a less defined sentence 
structure, with keywords appended to the end, while the fourth video mentions a specific place (maldives). Time-specific information is important for the second and third example, where details such as ``talking on walkie-talkie'' or ``playing billiards'' would be missed when looking at certain frames independently.


\begin{table}[h!]
\centering
\caption{\textbf{Dataset Statistics:} We train on a new dataset mined from the web called WebVid2M. Our dataset is an order of magnitude larger than existing video-text datasets in the number of videos and captions. HowTo100M (highlighted in blue) is a video dataset with noisy, weakly linked text supervision from ASR.}
\footnotesize
\begin{tabular}{@{}llrrrr@{}}
\toprule
\multicolumn{1}{l}{\textbf{dataset}} & \multicolumn{1}{c}{\textbf{domain}} & \multicolumn{1}{c}{\textbf{\begin{tabular}[c]{@{}c@{}}\#clips\end{tabular}}} & \multicolumn{1}{c}{\textbf{\begin{tabular}[c]{@{}c@{}}avg dur.\\ (secs)\end{tabular}}} & \multicolumn{1}{c}{\textbf{\#sent}} & \multicolumn{1}{c}{\textbf{\begin{tabular}[c]{@{}c@{}} time\\ (hrs)\end{tabular}}} \\ \midrule

MPII Cook~\cite{rohrbach2012database} & cooking & 44 & 600 & 6K & 8\\ 

TACos~\cite{regneri2013grounding} & cooking & 7K & 360 & 18K & 15.9  \\ 

DideMo~\cite{anne2017localizing} & flickr & 27K & 28 & 41K & 87\\ 
MSR-VTT~\cite{xu2016msr} & youtube & 10K & 15 & 200K & 40 \\ 
Charades~\cite{sigurdsson2016hollywood} & home & 10K & 30 & 16K & 82 \\ 
LSMDC15~\cite{rohrbach2017movie} & movies & 118K & 4.8 & 118K & 158\\ 
YouCook II~\cite{zhou2018towards} & cooking & 14K & 316 & 14K & 176 \\ 
ActivityNet~\cite{krishna2017dense} & youtube & 100K & 180 & 100K & 849 \\ 
CMD~\cite{bain2020condensed} & movies & 34K & 132 & 34K & 1.3K \\
\textbf{WebVid-2M} & open  & \textbf{2.5M} & 18 & \textbf{2.5M} & \textbf{13K} \\ 
\rowcolor{aliceblue}
HT100M~\cite{miech2019howto100m} & instruction & 136M & 4 & 136M &  134.5K \\

\bottomrule            
\end{tabular}
\label{tab:datastats}
\end{table}

We note that our video dataset is 10x smaller than HowTo100M in video duration and over 20x smaller in the number of paired clip-captions (Table \ref{tab:datastats}). Our dataset consists of manually generated captions, that are for the most part well formed sentences. In contrast, HowTo100M is generated from continuous narration with incomplete sentences that lack punctuation. The clip-text pairs are obtained from subtitles and may not be temporally aligned with the video they refer to, or indeed may not refer to the video at all~\cite{miech2019howto100m}. Our captions, on the other hand, are aligned with the video and describe visual content.

Moreover, there is no noise from imperfect ASR transcription and grammatical errors as is the case for HowTo100M. Our dataset also has longer captions on average (12 vs 4 words for HowTo) which are more diverse (Measure of Textual Lexical Diversity, MTLD~\cite{mccarthy2010mtld} = 203 vs 13.5).

\noindent\textbf{Image pretraining: Google Conceptual Captions~\cite{sharma2018conceptual}.}
This dataset consists of about 3.3M image and description pairs. Unlike the curated style of COCO images, Conceptual Captions (CC3M) images and their raw descriptions are harvested from the web, and therefore represent a wider variety of styles. The raw descriptions
are harvested from the Alt-text HTML attribute associated with web images.

\subsection{Downstream Datasets}
\label{subsec:datasets2}
We now describe the downstream text-video datasets that our model is evaluated on. \\
\noindent\textbf{MSR-VTT~\cite{xu2016msr}} contains
10K YouTube videos with 200K descriptions. Following other works~\cite{Liu19a}, we train on 9K train+val videos and report results on the 1K-A test set. \\
\noindent\textbf{MSVD~\cite{chen2011collecting}} consists of 80K English descriptions for 1,970 videos from YouTube, with each video containing 40 sentences each. We use the standard split of 1200, 100, and 670 videos for training, validation, and testing~\cite{patrick2020support,Liu19a}.

\noindent\textbf{DiDeMo~\cite{anne2017localizing}} contains 10K Flickr videos annotated with 40K sentences. Following \cite{lei2021less,Liu19a}, we evaluate paragraph-to-video retrieval, where all sentence descriptions for a video are concatenated into a single query. Since this dataset comes with localisation annotations (ground truth proposals), we report results with ground truth proposals (where only the localised moments in the video are concatenated and used in the retrieval set as done by~\cite{lei2021less}) as well as without (as done by~\cite{Liu19a}). 

\noindent\textbf{LSMDC~\cite{Rohrbach_2015_CVPR}} consists of 118,081 video clips sourced from 202 movies. The validation set contains 7,408 clips and evaluation is done on a test set of 1,000 videos from movies disjoint from the train and val sets. This follows the protocol outlined in~\cite{rohrbach2017movie}.

\noindent\textbf{Flickr30K~\cite{young-etal-2014-image}}. We also evaluate on a text-to-image retrieval benchmark to demonstrate the versatility of our model in that it can be used to achieve competitive performance in image settings as well as state-of-the art in video retrieval. The Flickr30K dataset contains 31,783 images with 5 captions per image. We follow the standard protocol of 1,000 images for validation, 1,000 images for testing and the remaining for training.

For downstream datasets with separate \texttt{val} and \texttt{test} splits, we train all models for 75 epochs and use the epoch with the lowest validation loss for reporting test results. For downstream datasets without a \texttt{val} set we report results at 50 epochs.

\subsection{Implementation Details}
\label{subsec:implementation}
All experiments are conducted with PyTorch~\cite{NEURIPS2019_9015}. Optimization is performed with Adam, using a learning rate of $1 \times 10^{-5}$, we use batch sizes of 16, 24, and 96 for 8, 4, and 1-frame inputs respectively. The temperature hyperparameter $\sigma$ for the loss defined in Eq.~\ref{eq:loss1} \&~\ref{eq:loss2} is set to 0.05. The default pretraining is WebVid-2M and CC3M.

For the visual encoder, all models have the following: $|\ell|=12$ attention blocks, patch size $P=16$, sequence dimension $D=768$, 12 heads and takes 4-frames as downstream input.

The text encoder of all models, unless specified otherwise, is instantiated as DistilBERT base-uncased~\cite{distilbert} pretrained on English Wikipedia and Toronto Book Corpus. The dimensionality of the common text-video space is set to 256. For visual augmentation, we randomly crop and horizontally flip during training, and center crop the maximal square crop at test time. All videos are resized to $224 \times 224$ as input. At test-time we compute clip-embeddings for the video with a stride of 2 seconds.
For paragraph-retrieval settings, we employ text augmentation during training by randomly sampling and concatenating a variable number of corresponding captions per video.

\noindent\textbf{Finetuning time.}
A large motivation for using pre-extracted expert models for video retrieval is to save computational cost. Finetuning our 4-frame model for 50 epochs on MSR-VTT takes 10 hours on 2 Quadro RTX 6000k GPUs (with 24GB RAM each), which is similar to other works using \textit{pre-extracted expert features}~\cite{patrick2020support}. This shows that our model is lightweight and can be finetuned end-to-end on the downstream video datasets quickly with sufficient pretraining (which is of one-time cost).
\subsection{Ablation Study}
\label{subsec:ablation}

\begin{table}
\centering
\caption{\textbf{Pretraining sources:} The effect of different pretraining sources. We use 4 frames per video in both pretraining and finetuning. Pretraining is performed for 1 full epoch only. Results are presented on the 1K-A MSR-VTT test set for text-video retrieval. \textbf{R@k:} Recall@K. \textbf{MedR:} Median Rank}
\resizebox{0.99\linewidth}{!}{
\begin{tabular}{lrrrrr}
\toprule
\textbf{Pre-training} & \textbf{\#pairs} & \textbf{R@1} & \textbf{R@10} & \textbf{MedR}    \\
\midrule
-                   & -                   & 5.6     & 22.3     &  55  \\
ImageNet            &                     & 15.2    & 54.4     &  9.0     \\
HowTo-17M subset     & 17.1M                  & 24.1   & 63.9     & 5.0     \\
CC3M                & 3.0M                  & 24.5    & 62.7     & 5.0     \\
WebVid2M            & 2.5M                  & 26.0   & 64.9      & 5.0 \\
\textbf{CC3M + WebVid2M}    & 5.5M                   & \textbf{27.3}    & \textbf{68.1}         & \textbf{4.0} \\ \bottomrule
\end{tabular}
}
\label{tab:pretraining}
\end{table}

In this section we study the effect of different pretraining strategies. In the Section~\ref{sec:arch_abl} of the Appendix, we provide architectural ablations on different temporal expansion methods, different visual backbones, different text backbones and the improvement when using our modified space-time attention block.

\noindent\textbf{Effect of pretraining.}
We compare performance on MSR-VTT with our model (i) trained from scratch, (ii) initialised with ImageNet weights and then 
finetuned, as well as (iii) initalised with ImageNet, and then pretrained on a number of different visual-text datasets before 
finetuning. For the video data, 4 frames are sampled at both pretraining and finetuning. Results on the MSR-VTT 1KA test set are shown in  Table~\ref{tab:pretraining}.
For HowTo100M, we pretrain on a random 17M subset due to computational constraints (the largest subset we could obtain at the time of writing) totalling 19K hours. To generate text-video pairs, we sample 5 contiguous speech-video pairs and concatenate them to form a longer video. This allows for robustness to the noisy alignment of speech and vision. We find that training on CC3M alone does reasonably well, outperforming the HowTo-17M subset. This demonstrates the benefit of our flexible encoder that can be cheaply trained on images and easily applied to videos. Training on WebVid2M also outperforms training on the HowTo17M subset, despite being much smaller, confirming that the HowTo100M dataset is noisy. The best performance is achieved by jointly training on both CC3M and WebVid2M, effectively exploiting image and video data. 

\subsection{Curriculum strategy}
\label{subsec:curriculumexp}
Next, we evaluate the ability of our curriculum schedule
to gradually learn the temporal dimension of videos
by increasing the input number of frames.
Table~\ref{tab:curric} summarises the results.
Here, we show performance when pretraining on WebVid2M and
finetuning on MSR-VTT.
We explore two types of expansion in time:
at pretraining and at finetuning stages.
First, we observe that a single frame is not sufficient
to capture the video content (18.8 R@1).
Performing the temporal expansion at pretraining stage
is better than doing so at finetuning (26.0 vs 24.9 R@1 with 4 frames).
Finally, we obtain similar performance (slightly better at R@5) at half the computational
cost in GPU hours by employing a curriculum strategy at pretraining (26.6 R@1).
For 8 frames, the curriculum is even more useful, as we start training on 1 frame and then move to 4 before finally moving to 8 frames. Here, we obtain similar or better performance than training on 8 frames from the start, with almost a third of the computational cost. This is to be expected, as fewer frames significantly reduces forward pass times and enables larger batch sizes.
Note that for a fair comparison, we allow the same number of training
iterations for each row in the table.
\begin{table}
\centering
\caption{\textbf{Effect of \#frames and curriculum learning:} The effect of a different number of input frames at pretraining and finetuning. $\Rightarrow$ indicates a within-dataset curriculum learning strategy. Results are presented on the 1K-A MSR-VTT test set for text-video retrieval. Pretraining here is done on WebVid2M only, with a total budget of one epoch through the entire dataset. \textbf{PTT:} total pretraining time in hours.}
\resizebox{0.99\linewidth}{!}{
\begin{tabular}{@{}ccrrrr@{}}
\toprule
\textbf{PT \#frames}          & \textbf{FT \#frames} & \textbf{R@1} & \textbf{R@10} & \textbf{MedR} & \textbf{PTT (hrs)} \\ \toprule
1                             & 1                    & 18.8         & 56.6          & 7.0 & 16.2          \\ \midrule
1                             & 4                    & 24.9         & 67.1          & 5.0 & 16.2           \\
4                             & 4                    & 26.0         & 64.9          & 5.0 & 45.6        \\
1$\Rightarrow$4               & 4                    & 26.6         & 65.5          & 5.0 & 22.1             \\ \midrule
8                             & 8                    & 25.4         & 67.3          & 4.0 & 98.0           \\
1$\Rightarrow$4$\Rightarrow$8 & 8                    & 27.4         & 67.3          & 4.0 & 36.0         \\ \bottomrule
\end{tabular}
}
\label{tab:curric}
\end{table}

\begin{figure}
    \centering
    \includegraphics[width=0.5\textwidth]{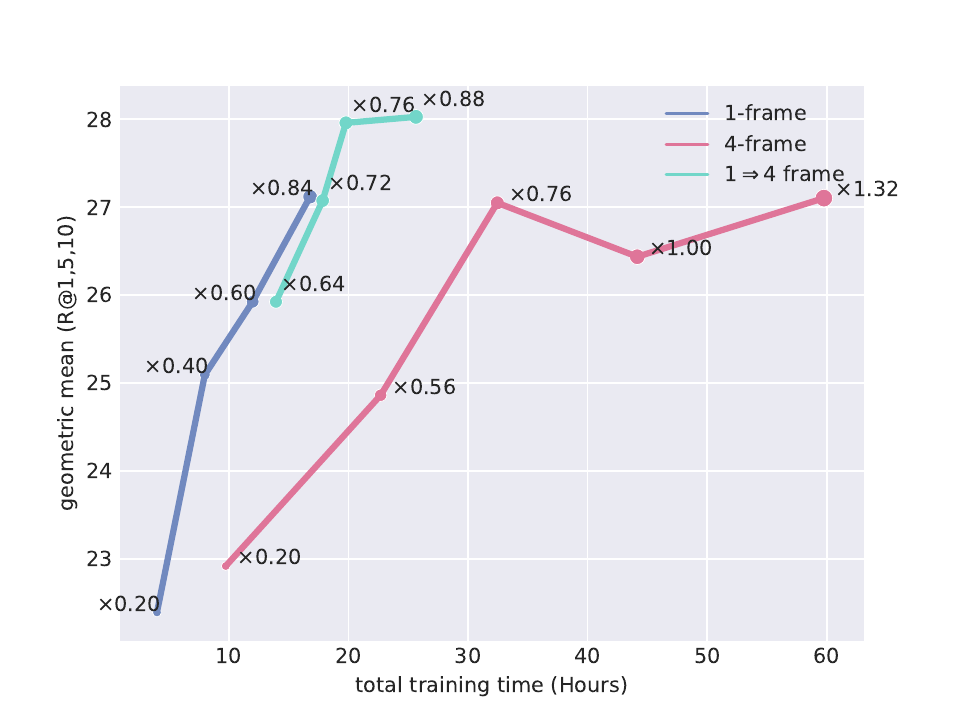}
    \caption{Plot showing the zero-shot performance (geometric mean of R@{1,5,10}) of various models on the MSR-VTT test set against their total training time in hours. $\Rightarrow$ denotes a curriculum learning strategy. $\times$ denotes the multiple of dataset epochs completed.}
    \label{fig:curric}
\end{figure}

We further analyse our proposed temporal curriculum strategy and its effects on training time and accuracy. Figure~\ref{fig:curric} shows the zero-shot results on MSR-VTT for various checkpoints with and without curriculum. It shows that our curriculum method yields a significant training speedup with a gain in accuracy. Shorter frame models are able to pass through more of the dataset in a shorter amount of time, which can lead to significant performance benefits in a constrained setting.

\begin{table*}[ht]
\setlength{\tabcolsep}{5.5pt}
\centering
\caption{\label{tab:msr-vtt-sota}Comparison to state-of-the-art results on MSR-VTT for text-to-video retrieval, 1k-A split. $\dagger$\textbf{E2E:} Works trained on pixels directly, without using pre-extracted expert features trained for other tasks. \textbf{Vis Enc. Init.:} Datasets used for pretraining visual encoders for tasks \textit{other than visual-text retrieval}, eg object classification. \textbf{Visual-Text PT:} Visual-text pretraining data. Rows highlighted in blue use additional modalities such as sound and speech from the MSR-VTT test videos. $\dagger$ Object, Motion, Face, Scene, Speech, OCR and Sound classification features.}
\begin{tabular}{@{}llllrrrrr@{}}
\toprule
\textbf{Method} & \textbf{E2E$\dagger$} & \textbf{Vis Enc. Init.} &\textbf{Visual-Text PT} & \textbf{\#pairs PT} & \textbf{R@1} & \textbf{R@5} & \textbf{R@10} & \textbf{MedR} \\ \midrule
JSFusion~\cite{yu2018joint} & $\checkmark$ & - & -  & - &10.2 & 31.2 & 43.2 & 13.0   \\
HT MIL-NCE~\cite{miech2019howto100m} &$\checkmark$ & - &HowTo100M & 136M & 14.9 & 40.2 & 52.8 & 9.0   \\
ActBERT~\cite{zhu2020actbert} & $\checkmark$ & VisGenome &HowTo100M & 136M & 16.3 & 42.8 & 56.9 & 10.0  \\
HERO~\cite{li2020hero} & $\checkmark$ & ImageNet, Kinetics &HowTo100M & 136M & 16.8 & 43.4 & 57.7 & -  \\
VidTranslate~\cite{korbar2020video} & $\checkmark$ & IG65M &HowTo100M & 136M & 14.7 & - & 52.8  \\
NoiseEst.~\cite{amrani2020noise} & \xmark & ImageNet, Kinetics  &HowTo100M & 136M & 17.4 & 41.6 &  53.6 & 8.0  \\
\rowcolor{aliceblue} CE~\cite{Liu19a} & \xmark& Numerous experts$\dagger$ &   -  & & 20.9 & 48.8 & 62.4 & 6.0\\
UniVL~\cite{luo2020univilm} & \xmark & - &HowTo100M & 136M & 21.2 & 49.6 & 63.1 & 6.0  \\ 
ClipBERT~\cite{lei2021less} & \checkmark & - &COCO, VisGenome & 5.6M & 22.0 & 46.8 & 59.9 & 6.0   \\ 
AVLnet~\cite{rouditchenko2020avlnet} & \xmark &ImageNet, Kinetics &HowTo100M & 136M & 27.1 & 55.6 & 66.6 & 4.0  \\
\rowcolor{aliceblue} MMT~\cite{gabeur2020multi} & \xmark&  Numerous experts$\dagger$ &HowTo100M & 136M & 26.6 & 57.1 & 69.6 & 4.0 \\
\rowcolor{aliceblue} T2VLAD~\cite{wang2021t2vlad} & \xmark&  Numerous experts$\dagger$ & - & & 29.5 & 59.0 & 70.1 & 4.0 \\
Support Set~\cite{patrick2020support} & \xmark & IG65M, ImageNet& - &- &
27.4 & 56.3 & 67.7 & 3.0 \\
Support Set~\cite{patrick2020support} & \xmark& IG65M, ImageNet&HowTo100M & 136M & 30.1 & 58.5 &69.3 & \textbf{3.0}  \\
\textbf{Ours} &  \checkmark & ImageNet & CC3M & 3M & 25.5  & 54.5  & 66.1  & 4.0 \\ 
\textbf{Ours} &  \checkmark & ImageNet  & CC3M,WV-2M & 5.5M & 
\textbf{31.0} & \textbf{59.5} & \textbf{70.5} & \textbf{3.0} \\ 
\textbf{Ours} &  \checkmark & ImageNet  & CC3M,WV-2M,COCO & 6.1M & \textbf{32.5} & \textbf{61.5} & \textbf{71.2} & \textbf{3.0} \\ 
\midrule
\textbf{Zero-shot} \\
\midrule
HT MIL-NCE~\cite{miech2019howto100m} &$\checkmark$ & - &HowTo100M & 136M & 7.5 & 21.2 & 29.6 & 38.0     \\
SupportSet~\cite{patrick2020support} &  & IG65M, ImageNet &HowTo100M & 136M & 8.7 & 23.0 & 31.1 & 31.0     \\
\textbf{Ours} & \checkmark & ImageNet & CC3M,WV-2M & 5.5M & \textbf{23.2} & \textbf{44.6} & \textbf{56.6} & \textbf{7.0} \\
\textbf{Ours} & \checkmark & ImageNet & CC3M,WV-2M,COCO & 6.1M & \textbf{24.7} & \textbf{46.9} & \textbf{57.2} & \textbf{7.0} \\
\bottomrule

\end{tabular}
\end{table*}

\noindent\textbf{Expansion of temporal embeddings.}
We experiment with both zero padding and interpolation, and find that our model is robust to the type of temporal expansion strategy. More detailed results are provided in the Appendix, Section~\ref{sec:temp_expansion}.

\subsection{Comparison to the State of the Art}
\label{subsec:sota}
Results on MSR-VTT can be seen in Table~\ref{tab:msr-vtt-sota}.
We outperform all previous works, including many that pretrain on HowTo100M which is an order of magnitude larger than our pretraining dataset both in the number of hours (135K vs 13K) and in the number of caption-clip pairs (136M vs 5.5M). We also note that we outperform works that extract expert features (CE uses 9 experts, MMT uses 7) including object, motion, face, scene, sound and speech embeddings. We even outperform/perform on par with Support Set~\cite{patrick2020support}, which uses expert features from a 34-layer, R(2+1)-D model pretrained on IG65M, concatenated with ImageNet ResNet152 features, after which they add a transformer network and train end-to-end on HowTo100M. 

We also report zero-shot results (Table~\ref{tab:msr-vtt-sota}) with no finetuning on MSR-VTT, outperforming both MIL-NCE and Support Set that trains on HowTo100M. This shows that our model is more generalisable, and can be used out of the box, and also perhaps that the domain of WebVid-2M is closer to that of MSR-VTT than HowTo100M. We will release the weights of our models publicly.

For both the zero-shot and finetuned setting we show that the addition of the COCO Captions image dataset further boosts our state-of-the-art MSR-VTT performance, indicating that the model is not yet saturated and additional pretraining dataset will lead to even better downstream performance.

For MSVD~\cite{chen2011collecting}, we outperform all previous methods (Table ~\ref{tab:msvd-sota}). In particular, we outperform Support Set~\cite{patrick2020support} even though they train on an order of magnitude more data.
\begin{table}
\centering
\caption{Text-to-video retrieval results on the MSVD~\cite{chen2011collecting} test set.}
\resizebox{0.99\linewidth}{!}{
\begin{tabular}{@{}lrrrr@{}}
\toprule
\textbf{Method}  & \textbf{R@1}  & \textbf{R@5}  & \textbf{R@10} & \textbf{MedR} \\ \midrule
VSE~\cite{kiros2014unifying}             & 12.3          & 30.1          & 42.3          & 14.0          \\
VSE++~\cite{faghri2017vse++}              & 15.4          & 39.6          & 53.0          & 9.0           \\
Multi. Cues~\cite{mithun2018learning}            & 20.3          & 47.8          & 61.1          & 6.0           \\
CE~\cite{Liu19a}                   & 19.8          & 49.0          & 63.8          & 6.0           \\
Support Set~\cite{patrick2020support}           & 23.0          & 52.8          & 65.8          & 5.0           \\
Support Set~\cite{patrick2020support} (HowTo PT)   & 28.4          & 60.0          & 72.9          & 4.0           \\
\textbf{Ours}    & \textbf{33.7} & \textbf{64.7} & \textbf{76.3} & \textbf{3.0}  \\ \bottomrule
\end{tabular}
}
\label{tab:msvd-sota}
\end{table}

Results on DiDeMo can be found in Table~\ref{tab:didemo-sota}. Note that on this dataset, our zero-shot performance is equivalent to CLIPBERT's results with finetuning, and after we finetune our model on the DiDeMo training set we get an additional 14.2\% boost in R@1.
\begin{table}
\centering
\caption{Text-to-video retrieval results on the DiDeMo test set. We show results with and without ground truth proposals (GT prop.) as well as with finetuning and without (zero-shot).}
\resizebox{0.99\linewidth}{!}{
\begin{tabular}{@{}llrrrr@{}}
\toprule
\textbf{Method} & \textbf{GT prop.} & \textbf{R@1}  & \textbf{R@5}  & \textbf{R@10} & \textbf{MedR} \\ \midrule
S2VT~\cite{venugopalan2014translating}            &          & 11.9          & 33.6          & -             & 13.0          \\   
FSE~\cite{zhang2018cross}             &          & 13.9          & 36.0          & -             & 11.0          \\
CE~\cite{Liu19a}&                    & 16.1          & 41.1          & -             & 8.3           \\
ClipBERT~\cite{lei2021less}        & \checkmark         & 20.4          & 44.5          & 56.7          & 7.0           \\
\textbf{Ours}   & \textbf{}          & \textbf{31.0} & \textbf{59.8} & \textbf{72.4} & \textbf{3.0}    \\
\textbf{Ours}   & \checkmark          & \textbf{34.6} & \textbf{65.0} & \textbf{74.7} & \textbf{3.0}  \\  
\midrule 
\textbf{Zero-shot} \\
\midrule 

\textbf{Ours}   &          & 21.1 & 46.0 & 56.2 & 7.0 \\
\textbf{Ours}   & \checkmark          & 20.2 & 46.4 & 58.5 &  7.0

\\\bottomrule
\end{tabular}
}
\vspace{-1em}
\label{tab:didemo-sota}
\end{table}

We demonstrate further state-of-the-art results on LSMDC text-to-video retrieval.  We outperform all previous methods, except for MMT in Median Rank, which pretrains on HowTo100M, a dataset consisting of over 100M clip-text pairs and contains multiple experts as well as audio modalities. Our model uses visual information alone.
\begin{table}
\centering
\caption{Text-to-video retrieval results on the LSMDC test set.}
\resizebox{0.99\linewidth}{!}{
\begin{tabular}{@{}lrrrr@{}}
\toprule
\textbf{Method}  & \textbf{R@1}  & \textbf{R@5}  & \textbf{R@10} & \textbf{MedR} \\ \midrule
JSFusion~\cite{yu2018joint}     & 9.1          & 21.2          & 34.1          & 36.0  \\
MEE~\cite{miech18learning}      & 9.3          & 25.1          & 33.4          & 27.0  \\
CE~\cite{Liu19a}                & 11.2          & 26.9          & 34.8          & 25.3  \\
MMT (HowTo100M)~\cite{gabeur2020multi}                & 12.9          & 29.9          & 40.1          & \textbf{19.3}  \\
\textbf{Ours}                   & \textbf{15.0} & \textbf{30.8} & \textbf{40.3} & 20.0  \\ \bottomrule
\end{tabular}
}
\label{tab:lsmdc-sota}
\end{table}
\label{subsec:qualitative}

To demonstrate the effectiveness of our model for downstream video and image tasks, we additionally report results on Flickr30K the image retrieval dataset in Table~\ref{tab:flickr}. Unlike other works~\cite{lee2018stacked,chen2020imram,diao2021similarity} which utilise high resolution regions extracted using a Faster-RCNN detector, our model is single stage and does not require any object detections. We compare to works with a similar number of training image-text pairs, and find that our model is comparable. We also note that training on WebVid2M provides a sizeable boost (5\% improvement in R@1). Note that there are other recent text-image works such as UNITER~\cite{chen2020uniter} and OSCAR~\cite{li2020oscar}, however these are trained on almost twice the number of samples. Recent works scale this up even further to billions of samples (ALIGN~\cite{jia2021scaling}). 

\begin{table}
\centering
\caption{Text-to-\textbf{image} retrieval results on the Flickr30K test set. ++ indicates additional datasets: COCO Captions, SBU Captions. VisGenObjects denotes Visual Genome object bounding box annotations used to pretrain an FRCNN object feature extractor.} 
\resizebox{0.99\linewidth}{!}{
\begin{tabular}{@{}lrrrr@{}}
\toprule
\textbf{Method} & \multicolumn{1}{l}{\textbf{Vis PT. size}} & \multicolumn{1}{l}{\textbf{R@1}} & \multicolumn{1}{l}{\textbf{R@5}} & \multicolumn{1}{l}{\textbf{R@10}} \\ \midrule
SCANM~\cite{lee2018stacked}           & VisGenObj (3.8M)                             & 48.6                             & 77.7                             & 85.2                              \\
IMRAM~\cite{chen2020imram}           & VisGenObj (3.8M)                             & 53.9                             & 79.4                             & 87.2                              \\
SGRAF~\cite{diao2021similarity}           & VisGenObj (3.8M)                             & 58.5                             & 83.0                             & 88.8                              \\
Ours            & CC (3.0M)                                 &               54.2   &                             83.2     &        89.8                           \\
Ours            & CC,WV-2M (5.5M)                      &                61.0           &                      87.5 &             92.7                      \\ \bottomrule
\end{tabular}
}
\label{tab:flickr}
\end{table}


\section{Extension: Scaling up Further}

To investigate the effects of downstream performance on additional pretraining datasets and increased scale, we train models on the following datasets:

\noindent\textbf{WebVid-10M:} An extension to our WebVid-2M dataset, we increase the size of the dataset fourfold to 10 million text-video pairs, following the same data collection protocol. The captions and video url's can also be found at \url{https://m-bain.github.io/webvid-dataset/}.

\noindent\textbf{Conceptual-Captions 12M~\cite{changpinyo2021cc12m}:} A dataset comprising of 12 million captioned images, intended for large-scale vision language pre-training. It is larger and more diverse than the Conceptual Captions (CC3M), albeit with noisier captions.

\noindent\textbf{COCO Captions~\cite{chen2015microsoft}:} A smaller dataset of 113.3k images with five captions per image, resulting in a total of 567k image-text pairs.

\begin{table}

\centering
\caption{\textbf{Pretraining sources extended:} The effect of different  other pretraining sources. We use 4 frames per video when finetuning. Results are presented on the 1K-A MSR-VTT test set for text-video retrieval.}
\resizebox{0.99\linewidth}{!}{
\begin{tabular}{lrrrrrr}
\toprule
\textbf{Pre-training} & \textbf{\#pairs} & \textbf{R@1} & \textbf{R@5} & \textbf{R@10} & \textbf{MedR}    \\
\midrule
COCO     & 0.6M   & 27.2 &  56.1 & 67.5            & 4.0 \\
  WV-2M                 & 2.5M   & 27.5 & 56.6  & 67.6            & 4.0 \\
  WV-10M                 & 10M   & 28.9 & 57.2  & 68.6            & 4.0 \\
  \midrule
 CC3M, WV2M          & 5.0M   & 31.0 & 59.5  & 70.5       & 3.0 \\
 CC3M, WV2M, COCO   & 5.6M   & 32.5 & 61.5 & 71.2        & 3.0 \\
 CC3M, WV10M         & 13.0M  & 33.4 & 59.2 & 70.7        & 3.0 \\
 CC3M, CC12M, WV10M        & 25.0M  & 34.0 & 61.4 & 73.1 & 3.0 \\ \bottomrule
\end{tabular}
}
\label{tab:mo_pretraining}
\end{table}

Downstream performance of these additional datasets can be found in Table~\ref{tab:mo_pretraining}. We find that restricting the model's pretraining to only a small number of text-image pairs (COCO Captions) expectedly performs worse on downstream data, but still achieves competitive results. Thereby demonstrating the strength of our proposed method and that reasonable performance can be achieved on downstream video data with image pretraining alone.

Increasing the number of pretraining pairs consistently improves downstream performance, albeit with diminishing returns. It appears to be more efficient to add smaller datasets from diverse sources rather than add an increasingly larger dataset from a single source, shown by the boost of adding COCO captions (567k pairs) to the CC3M+WV2M pretraining compared to adding an extra 7.5 million pairs (WebVid10M) of that same source of data.
\section{Conclusion}
\label{sec:conc}
To conclude, we introduce a dual encoder model for end-to-end training of text-video retrieval, designed to take advantage of both large-scale image and video captioning datasets. Our model achieves state-of-the-art performance on a number of downstream benchmarks, however we note that the performance of our model is not saturated yet, and performance could be further improved by training on the full HowTo100M dataset, larger weakly paired image datasets such as Google3BN~\cite{jia2021scaling}, as well as multi-dataset combinations thereof.

\noindent \textbf{Acknowledgements.}
The authors would like to thank Samuel Albanie for his useful feedback. We are grateful for funding from a Royal Society Research Professorship,
EPSRC Programme Grant VisualAI EP/T028572/1,
and a Google PhD Fellowship.


\clearpage
\appendix
\addcontentsline{toc}{section}{Appendix} 
\part{Appendix} 
{\hypersetup{linkcolor=black} \parttoc} 

\section{Additional Benchmark Results}




ActivityNet Captions~\cite{krishna2017dense} contains 20K YouTube videos focused on actions, annotated with 100K sentences. The training set consists of 10K videos, and we use the `val1' set of 4.9K videos to report results. At test time we use paragraph-to-video retrieval as is standard protocol set by other works, where the segment descriptions are concatenated to give a video-level description. We compare to prior work in Table~\ref{tab:activitynet-sota} and achieve comparable results to the state of the art by using much less training data.
\begin{table}[h]
\centering
\caption{Text-to-video retrieval results on the ActivityNet val1k set. \textbf{R@k:} Recall@K. \textbf{MedR:} Median Rank.}
\resizebox{0.99\linewidth}{!}{
\begin{tabular}{lllrrr}
\toprule
\textbf{Method} & E2E        & VT PT     & \textbf{R@1}  & \textbf{R@5} &  \textbf{MedR} \\ \midrule
FSE             &            &           & 18.2          & 44.8         & 8.3           \\
CE~\cite{Liu19a}                 &            &           & 18.2          & 47.7         & 13.0          \\
CLIPBERT        &\checkmark  &           & 21.3          & 49.0         & 6.0           \\
MMT             &            &           & 22.7          & 54.2         & 5.0             \\
SupportSet~\cite{patrick2020support}      &            &           & 26.8          & 58.1         & \textbf{3.0}             \\
MMT~\cite{gabeur2020multi}          &            & HowTo & 28.7          & {61.4}                & \textbf{3.0}            \\
SupportSet~\cite{patrick2020support}      &            & HowTo & \textbf{29.2} & \textbf{61.6}       & \textbf{3.0}\\
\textbf{Ours}   & \checkmark & CC,WebVid-2M          & 28.8     & 60.9    & \textbf{3.0} \\
\bottomrule
\end{tabular}
}
\label{tab:activitynet-sota}
\end{table}



\section{Architectural Details}

\subsection{Video Encoder}
\begin{figure}
    \centering
    \includegraphics[width=0.3\textwidth]{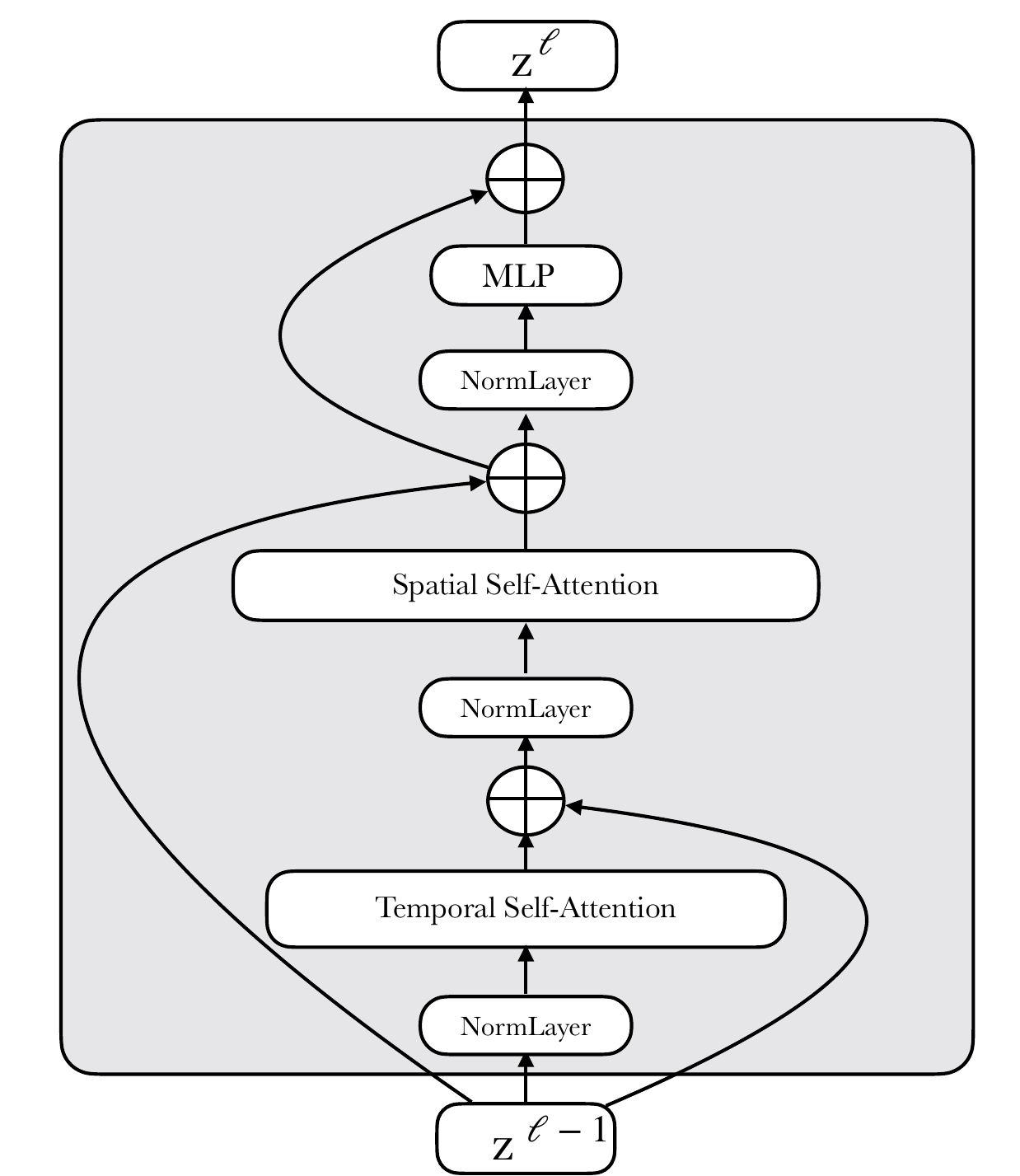}
    \caption{Detailed diagram of the space-time self attention block.}
    \label{fig:atten_block_indepth}
\end{figure}

The video encoder is composed of: (i) the patch embedding layer; (ii) learnable positional space, time and [CLS] embeddings; and (iii) a stack of $|\ell| = 12$ space-time attention blocks
\begin{enumerate}
    \item The patch embedding layer is implemented as a 2D convolutional layer with a kernel and stride size equivalent to the target patch size $P = 16$, and $d = 768$ output channels (the chosen embedding dimensionality of the video encoder).
    \item The positional space and time embeddings are instantiated with shape $M \times d$ and $N \times d$ respectively, where $M$ is the maximum number of input video frames and $N$ is the maximum number of non-overlapping patches of size $P$ within a frame (196 for a video resolution of $224\times224$). The [CLS] embedding is instantiated with shape $1 \times d$.
    \item Each space-time attention block consists of norm layers, temporal and spatial self-attention layers, and an MLP. The order and connections of these layers is shown in Figure~\ref{fig:atten_block_indepth}.
\end{enumerate}

\subsection{Text Encoder}

Our text encoder is instantiated as \texttt{distilbert-base-uncased}~\cite{distilbert}. Distilbert follows the same general architecture as BERT~\cite{devlin2019bert}, but with the number of layers reduced by a factor of 2 and the token-type embeddings and the pooler removed. We use the HuggingFace\footnote{\url{https://huggingface.co/}} transformers library implementation.

\section{Architectural Ablations}
\label{sec:arch_abl}
\subsection{Video Backbone}
We investigate the effects of using different video backbone architectures (Table~\ref{tab:vidbackbone}) and find that the space-time transformer encoder leads to large improvements in performance on MSR-VTT when compared to ResNets and 3D variants thereof.

During testing, all frame-variants see an equal number of frames, since the video embeddings are averaged over multiple strides.

For the video backbone ablation, we fix the text backbone to \texttt{distilbert-base-uncased}. For the text backbone ablation, we fix the video backone to the base space-time transformer with an input resolution of 224 and a patch size $P=16$.
\begin{table}[h]
\centering
\caption{\textbf{Video backbone.} Text-to-video retrieval results on MSR-VTT test set with different video backbones. All models were pretrained on WebVid-2M and finetuned on MSR-VTT train set. 4 frames were given as input, except for the ResNet-101 which only supports image (1-frame) inputs. The text backbone is fixed to distilbert-base-uncased.}
\begin{tabular}{@{}lrrrr@{}}
\toprule
\textbf{Video Backbone}   & \textbf{\#params} & \textbf{R@1} & \textbf{R@10} & \textbf{MedR} \\ \midrule
ResNet-101                &     45M &         11.5 &          44.1 & 14.5          \\
S3D-G                     &     76M &          3.6 &          20.4 & 59.5          \\
R(3D)-101                 &     85M &          9.3 &         38.3 &  20.0       \\
S-Tformer $224_{16}$ B    &    114M &\textbf{26.8} & \textbf{68.2} & \textbf{4.0}        \\
\bottomrule
\end{tabular}
\label{tab:vidbackbone}
\end{table}

\subsection{Text Backbone}
The choice of text backbone has a significant impact on downstream performance (Table~\ref{tab:txtbackbone}), with the t5 models performing significantly worse with more or similar numbers of parameters. DistilBERT and normal BERT achieve similar performance, with DistilBERT having far fewer parameters, therefore we chose to use DistilBERT in our work for efficiency.
\begin{table}
\centering
\caption{\textbf{Text backbone.} Text-to-video retrieval results on MSR-VTT test set with different text backbones. All models were pretrained on WebVid-2M and finetuned on MSR-VTT train set. The video backbone is fixed to the base space-time transformer with an input resolution of 224 and a patch size $P=16$.}
\resizebox{0.99\linewidth}{!}{
\begin{tabular}{@{}lrrrr@{}}
\toprule
\textbf{Text Backbone} & \textbf{\#params} & \textbf{R@1} & \textbf{R@10} & \textbf{MedR} \\ \midrule
t5-small               &    60.5M &         15.1 &          51.4 &          10.0     \\
t5-base                &   222.9M &         24.0 &          62.8 &          6.0     \\
distilbert-base-uncased  &    66.4M & 26.8 & \textbf{68.2} & \textbf{4.0}  \\
bert-base-uncased        &   109.5M &  \textbf{27.5} &  67.3 & \textbf{4.0} \\

\bottomrule
\end{tabular}
}
\label{tab:txtbackbone}
\end{table}

\subsection{Space-Time Attention}
\label{sec:spacetime_block}

\begin{figure}[h]
    \centering
    \includegraphics[width=0.5\textwidth]{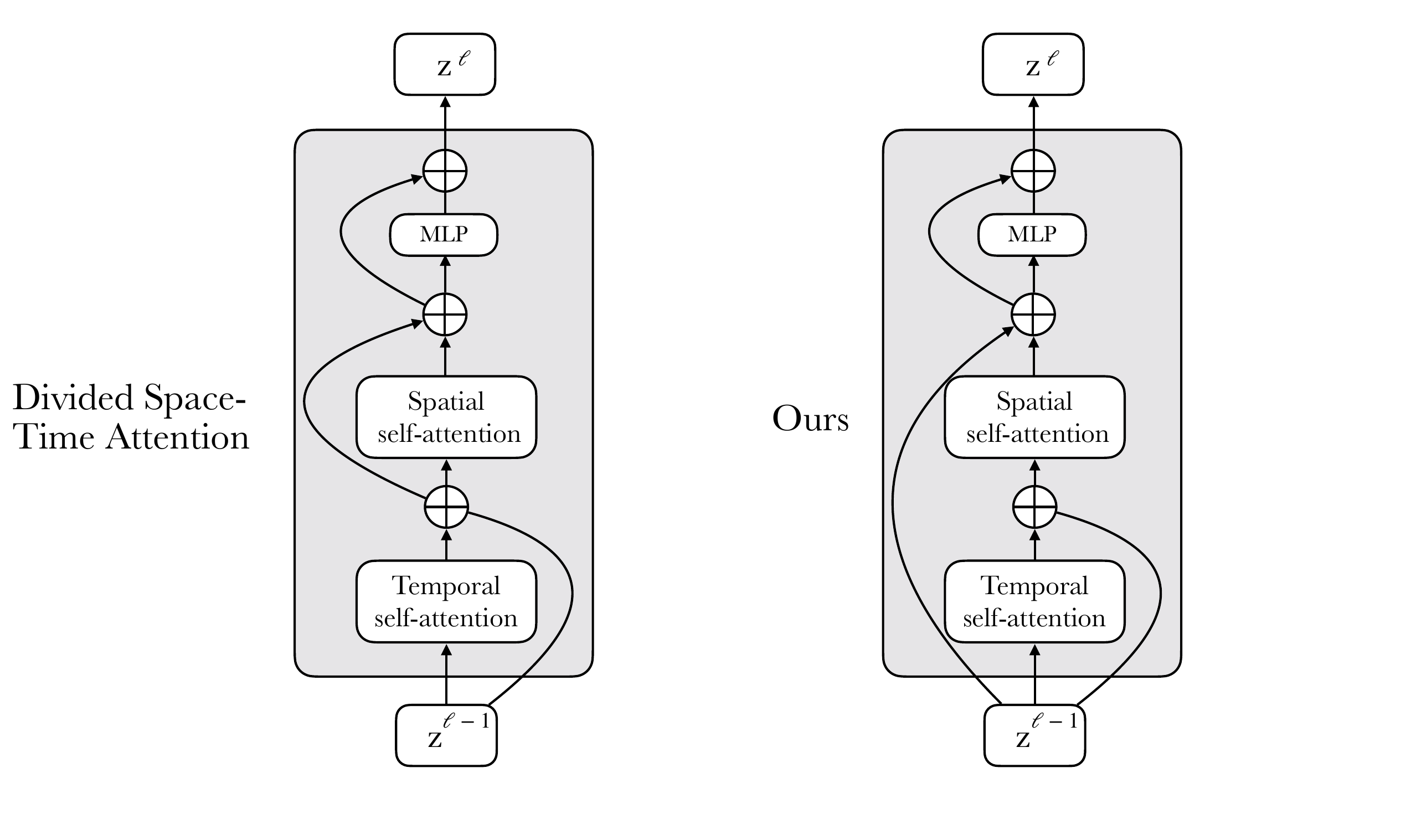}
    \caption{\textbf{Attention block:} The original divided block used in the Timesformer~\cite{bertasius2021spacetime} architecture (left) compared to ours (right). We find that this minor modification of the input residual connection trains more quickly and is more stable than the original.}
    \label{fig:spacetimeattn}
\end{figure}

\noindent\textbf{Space-time attention.} Our modified space-time attention block, shown in Fig. \ref{fig:spacetimeattn}, improves retrieval performance, as show in Table~\ref{tab:pret_attention}. We compare both variants during pretraining on WebVid-2M by reporting zero-shot results on MSR-VTT. We find once again that our modification leads to modest performance gains.

\begin{table}
\centering
\caption{\textbf{Space-time attention method:} Zero-shot results are presented on 1K-A MSR-VTT test set for text-video retrieval. The models were trained on WebVid-2M.}
\begin{tabular}{lrrrrr}
\toprule
\textbf{Attention Method} & \textbf{R@1} & \textbf{R@10} & \textbf{MedR}    \\
\midrule
Divided Space-Time~\cite{lei2021less} & 13.0    & 40.2      & 18.0  \\
Ours          & 14.6      & 42.7      & 16.0  \\
\bottomrule
\end{tabular}
\label{tab:pret_attention}
\end{table}

\subsection{Temporal Expansion}
\label{sec:temp_expansion}
\begin{table}[ht]
\centering
\caption{\textbf{Temporal expansion method}. The effect of different expansion methods increasing the input number of frames from 4$\Rightarrow$8. Results are presented on 1K-A MSR-VTT test set for text-video retrieval. The models were pre-trained on CC3M \& WebVid-2M and finetuned on MSR-VTT train set.}
\begin{tabular}{lrrrrr}
\toprule
\textbf{Method} & \textbf{R@1} & \textbf{R@10} & \textbf{MedR}    \\
\midrule
Zero-pad           & \textbf{30.7}       & 68.3      & 4.0   \\
Nearest Neighbour            & 29.4       & 69.5      & 4.0   \\
Bilinear           & 28.3    & \textbf{69.9}      & 4.0 \\
\bottomrule
\end{tabular}
\label{tab:temporalexp}
\end{table}
We explore 3 different methods for expanding temporal positional embeddings (zero-padding and two interpolation methods), and observe robustness to all 3 (see Table \ref{tab:temporalexp}).

\begin{figure*}[t]
\captionsetup[subfigure]{labelformat=empty}
     \centering
     \begin{subfigure}[t]{0.21\textwidth}
         \centering
         \includegraphics[width=\textwidth]{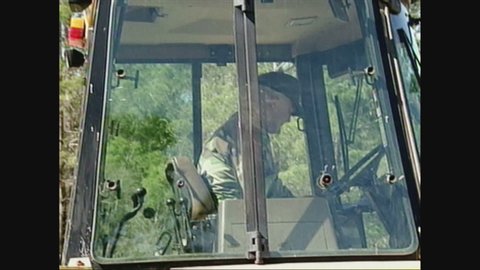}
         \caption{1990s: man driving excavator, rotates seat, opens windows in cab. hand presses lever.}
     \end{subfigure} \hfill
     \begin{subfigure}[t]{0.21\textwidth}
         \centering
         \includegraphics[width=\textwidth]{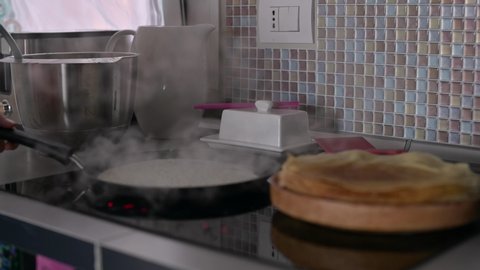}
         \caption{\footnotesize{Frying pancakes in the kitchen at home. a woman is cooking traditional russian pancakes. modern kitchen, skillet and batter.}}
         \label{fig:threesinx1}
     \end{subfigure} \hfill
     \begin{subfigure}[t]{0.21\textwidth}
         \centering
         \includegraphics[width=\textwidth]{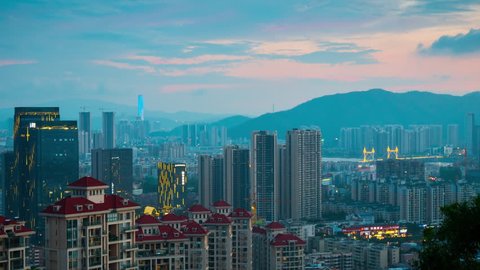}
         \caption{Twilight zhuhai famous mountain park top cityscape aerial panorama 4k timelapse china}
     \end{subfigure} \hfill
          \begin{subfigure}[t]{0.21\textwidth}
         \centering
         \includegraphics[width=\textwidth]{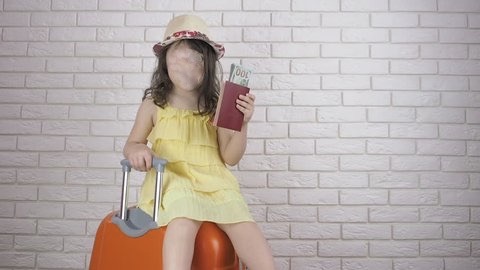}
         \caption{A child with a suitcase. a happy little girl sits on a suitcase with a passport and money.}
     \end{subfigure} \hfill 
     \begin{subfigure}[t]{0.21\textwidth}
         \centering
         \includegraphics[width=\textwidth]{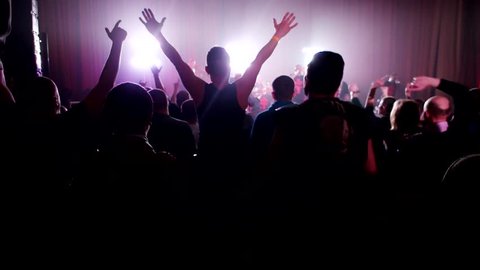}
         \caption{\tiny{Kherson, ukraine - 20 may 2016: open, free, rock music festival crowd partying at a rock concert. hands up, people, fans cheering clapping applauding in kherson, ukraine - 20 may 2016. band performing'}}
         \label{fig:threesinx2}
     \end{subfigure} \hfill
     \begin{subfigure}[t]{0.21\textwidth}
         \centering
         \includegraphics[width=\textwidth]{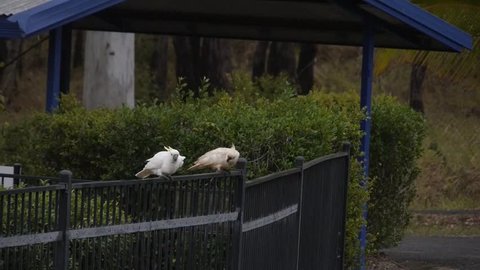}
         \caption{Cockatoos on the fence}
     \end{subfigure} \hfill
          \begin{subfigure}[t]{0.21\textwidth}
         \centering
         \includegraphics[width=\textwidth]{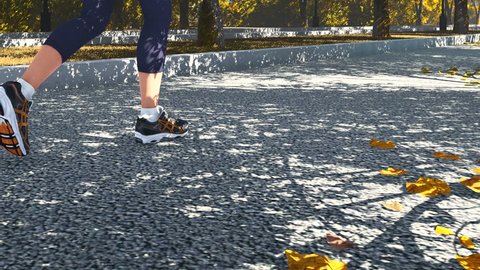}
         \caption{Runners feet in a sneakers close up. realistic three dimensional animation.}
     \end{subfigure} \hfill
     \begin{subfigure}[t]{0.21\textwidth}
         \centering
         \includegraphics[width=\textwidth]{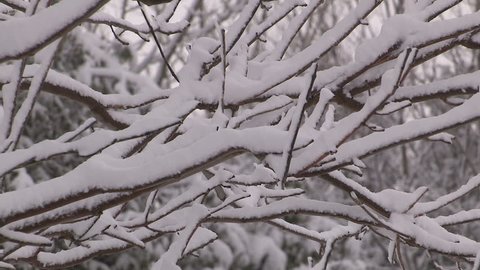}
         \caption{Ontario, canada january 2014 heavy pretty snow on tree branches}
         \label{fig:threesinx3}
     \end{subfigure} \hfill
        \caption{\textbf{WebVid-2M dataset examples:} We provide additional
        examples from our dataset by showing video-text pairs, using video thumbnails.}
        \label{fig:dataset_qual}
\end{figure*}

\begin{figure*}
    \centering
    \includegraphics[width=0.9\textwidth]{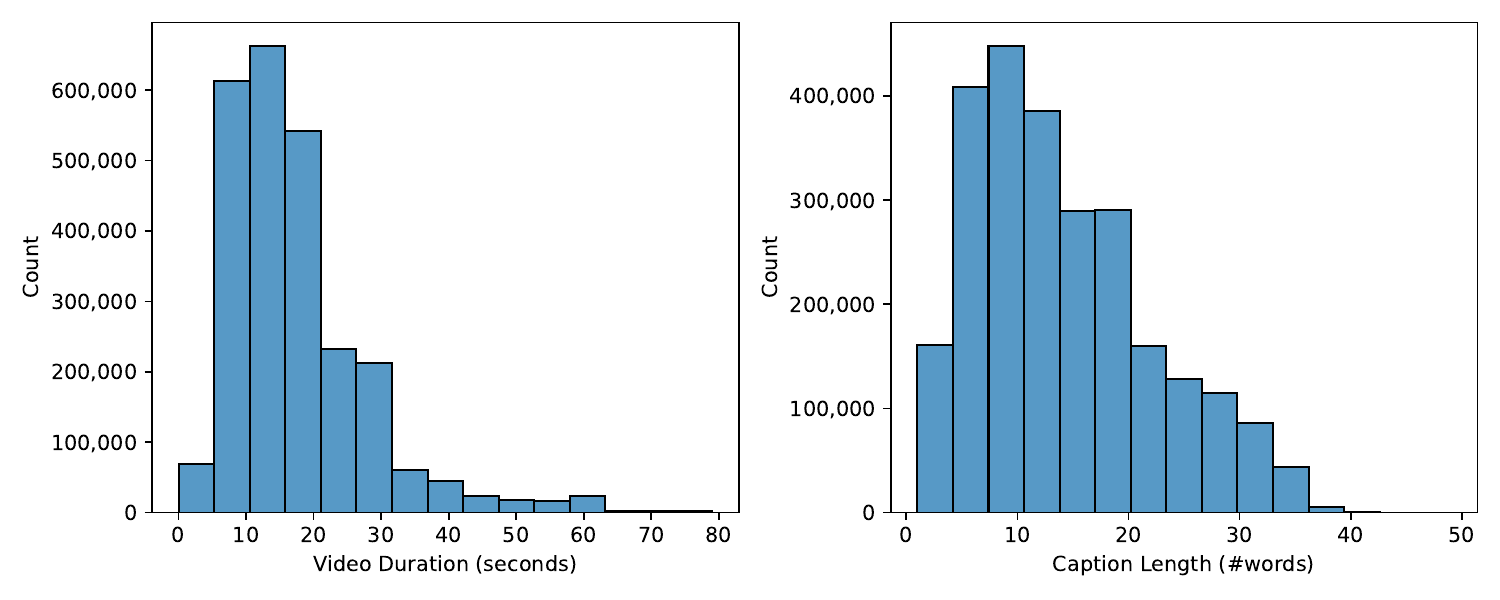}
    \caption{\textbf{WebVid-2M dataset statistics:} We report the histogram of video duration in seconds \textbf{(top)} and
    the histogram of caption length in words \textbf{(bottom)}.}
    \label{fig:dataset_hist}
    \vspace{-1em}
\end{figure*}

\section{WebVid-2M Dataset Details}
In this section, we show further details of the new WebVid-2M dataset. More qualitative examples of video-text pairs can be found in Figure~\ref{fig:dataset_qual} and histograms of caption lengths and video durations can be found in Figure~\ref{fig:dataset_hist}. Note that 275,000 videos are longer than 30 seconds, providing many examples of videos which can be used for training long-range video models.


\section{WebVid-10M Extension}
To facilitate further text-video pre-training, we extend the WebVid dataset fourfold to 10 million text-video pairs, following the same data collection protocol.

\begin{table}[h!]
\centering
\caption{\textbf{WebVid-10M:}}
\footnotesize
\begin{tabular}{@{}lrrrr@{}}
\toprule
\multicolumn{1}{l}{\textbf{dataset}} & \multicolumn{1}{c}{\textbf{\begin{tabular}[c]{@{}c@{}}\#clips\end{tabular}}} & \multicolumn{1}{c}{\textbf{\begin{tabular}[c]{@{}c@{}}avg dur.\\ (secs)\end{tabular}}} & \multicolumn{1}{c}{\textbf{\#sent}} & \multicolumn{1}{c}{\textbf{\begin{tabular}[c]{@{}c@{}} time\\ (hrs)\end{tabular}}} \\ \midrule

\textbf{WebVid-2M}  & 2.5M & 18 & 2.5M & 13K \\ 
\textbf{WebVid-10M}  & 10M & 18 & 10M & 52K \\ 

\bottomrule            
\end{tabular}
\label{tab:webvid_ext}
\end{table}
{\small\bibliographystyle{ieee_fullname}\bibliography{shortstrings,vgg_local,references}}

\begin{thebibliography}{10}\itemsep=-1pt

\bibitem{alayrac2020self}
Jean-Baptiste Alayrac, Adri{\`a} Recasens, Rosalia Schneider, Relja
  Arandjelovi{\'c}, Jason Ramapuram, Jeffrey De~Fauw, Lucas Smaira, Sander
  Dieleman, and Andrew Zisserman.
\newblock Self-supervised multimodal versatile networks.
\newblock In {\em NeurIPS}, 2020.

\bibitem{amrani2020noise}
Elad Amrani, Rami Ben~Ari, Daniel Rotman, and Alex Bronstein.
\newblock Noise estimation using density estimation for self-supervised
  multimodal learning.
\newblock {\em arXiv preprint arXiv:2003.03186}, 2020.

\bibitem{anne2017localizing}
Lisa Anne~Hendricks, Oliver Wang, Eli Shechtman, Josef Sivic, Trevor Darrell,
  and Bryan Russell.
\newblock Localizing moments in video with natural language.
\newblock In {\em ICCV}, 2017.

\bibitem{antol2015vqa}
Stanislaw Antol, Aishwarya Agrawal, Jiasen Lu, Margaret Mitchell, Dhruv Batra,
  C~Lawrence Zitnick, and Devi Parikh.
\newblock Vqa: Visual question answering.
\newblock In {\em ICCV}, 2015.

\bibitem{bain2020condensed}
Max Bain, Arsha Nagrani, Andrew Brown, and Andrew Zisserman.
\newblock Condensed movies: Story based retrieval with contextual embeddings.
\newblock In {\em ACCV}, 2020.

\bibitem{bertasius2021spacetime}
Gedas Bertasius, Heng Wang, and Lorenzo Torresani.
\newblock Is space-time attention all you need for video understanding?
\newblock {\em arXiv:2102.05095}, 2021.

\bibitem{carion2020endtoend}
Nicolas Carion, Francisco Massa, Gabriel Synnaeve, Nicolas Usunier, Alexander
  Kirillov, and Sergey Zagoruyko.
\newblock End-to-end object detection with transformers.
\newblock In {\em ECCV}, 2020.

\bibitem{Carreira2017}
Jo{\~{a}}o Carreira and Andrew Zisserman.
\newblock Quo vadis, action recognition? {A} new model and the {Kinetics}
  dataset.
\newblock In {\em CVPR}, 2017.

\bibitem{changpinyo2021cc12m}
Soravit Changpinyo, Piyush Sharma, Nan Ding, and Radu Soricut.
\newblock {Conceptual 12M}: Pushing web-scale image-text pre-training to
  recognize long-tail visual concepts.
\newblock In {\em CVPR}, 2021.

\bibitem{chen2011collecting}
David Chen and William~B Dolan.
\newblock Collecting highly parallel data for paraphrase evaluation.
\newblock In {\em Proceedings of the 49th Annual Meeting of the Association for
  Computational Linguistics: Human Language Technologies}, pages 190--200,
  2011.

\bibitem{chen2020imram}
Hui Chen, Guiguang Ding, Xudong Liu, Zijia Lin, Ji Liu, and Jungong Han.
\newblock {IMRAM}: Iterative matching with recurrent attention memory for
  cross-modal image-text retrieval, 2020.

\bibitem{chen2015microsoft}
Xinlei Chen, Hao Fang, Tsung-Yi Lin, Ramakrishna Vedantam, Saurabh Gupta, Piotr
  Dollar, and C.~Lawrence Zitnick.
\newblock Microsoft coco captions: Data collection and evaluation server, 2015.

\bibitem{chen20182}
Yunpeng Chen, Yannis Kalantidis, Jianshu Li, Shuicheng Yan, and Jiashi Feng.
\newblock A2-nets: Double attention networks.
\newblock {\em arXiv preprint arXiv:1810.11579}, 2018.

\bibitem{chen2020uniter}
Yen-Chun Chen, Linjie Li, Licheng Yu, Ahmed~El Kholy, Faisal Ahmed, Zhe Gan, Yu
  Cheng, and Jingjing Liu.
\newblock {UNITER}: Universal image-text representation learning, 2020.

\bibitem{child2019generating}
Rewon Child, Scott Gray, Alec Radford, and Ilya Sutskever.
\newblock Generating long sequences with sparse transformers.
\newblock {\em arXiv preprint arXiv:1904.10509}, 2019.

\bibitem{cordonnier2019relationship}
Jean-Baptiste Cordonnier, Andreas Loukas, and Martin Jaggi.
\newblock On the relationship between self-attention and convolutional layers.
\newblock {\em arXiv preprint arXiv:1911.03584}, 2019.

\bibitem{devlin2019bert}
J. Devlin, Ming-Wei Chang, Kenton Lee, and Kristina Toutanova.
\newblock {BERT}: Pre-training of deep bidirectional transformers for language
  understanding.
\newblock In {\em NAACL-HLT}, 2019.

\bibitem{diao2021similarity}
Haiwen Diao, Ying Zhang, Lin Ma, and Huchuan Lu.
\newblock Similarity reasoning and filtration for image-text matching, 2021.

\bibitem{dosovitskiy2021an}
Alexey Dosovitskiy, Lucas Beyer, Alexander Kolesnikov, Dirk Weissenborn,
  Xiaohua Zhai, Thomas Unterthiner, Mostafa Dehghani, Matthias Minderer, Georg
  Heigold, Sylvain Gelly, Jakob Uszkoreit, and Neil Houlsby.
\newblock An image is worth 16x16 words: Transformers for image recognition at
  scale.
\newblock In {\em ICLR}, 2021.

\bibitem{faghri2017vse++}
Fartash Faghri, David~J Fleet, Jamie~Ryan Kiros, and Sanja Fidler.
\newblock Vse++: Improving visual-semantic embeddings with hard negatives.
\newblock {\em arXiv preprint arXiv:1707.05612}, 2017.

\bibitem{gabeur2020multi}
Valentin Gabeur, Chen Sun, Karteek Alahari, and Cordelia Schmid.
\newblock Multi-modal transformer for video retrieval.
\newblock In {\em ECCV}, 2020.

\bibitem{girdhar2017actionvlad}
Rohit Girdhar, Deva Ramanan, Abhinav Gupta, Josef Sivic, and Bryan Russell.
\newblock Actionvlad: Learning spatio-temporal aggregation for action
  classification.
\newblock In {\em CVPR}, 2017.

\bibitem{HaraCVPR2018}
Kensho Hara, Hirokatsu Kataoka, and Yutaka Satoh.
\newblock Can spatiotemporal {3D} {CNNs} retrace the history of {2D} {CNNs} and
  {ImageNet}?
\newblock In {\em CVPR}, 2018.

\bibitem{hu2017relation}
Han Hu, Jiayuan Gu, Zheng Zhang, Jifeng Dai, and Yichen Wei.
\newblock Relation networks for object detection.
\newblock In {\em CVPR}, 2018.

\bibitem{jia2021scaling}
Chao Jia, Yinfei Yang, Ye Xia, Yi-Ting Chen, Zarana Parekh, Hieu Pham, Quoc~V
  Le, Yunhsuan Sung, Zhen Li, and Tom Duerig.
\newblock Scaling up visual and vision-language representation learning with
  noisy text supervision.
\newblock {\em arXiv preprint arXiv:2102.05918}, 2021.

\bibitem{Kinetics}
Will Kay, Joao Carreira, Karen Simonyan, Brian Zhang, Chloe Hillier, Sudheendra
  Vijayanarasimhan, Fabio Viola, Tim Green, Trevor Back, Paul Natsev, Mustafa
  Suleyman, and Andrew Zisserman.
\newblock The {Kinetics} human action video dataset.
\newblock {\em CoRR}, abs/1705.06950, 2017.

\bibitem{kiros2014unifying}
Ryan Kiros, Ruslan Salakhutdinov, and Richard~S Zemel.
\newblock Unifying visual-semantic embeddings with multimodal neural language
  models.
\newblock {\em arXiv preprint arXiv:1411.2539}, 2014.

\bibitem{korbar2020video}
Bruno Korbar, Fabio Petroni, Rohit Girdhar, and Lorenzo Torresani.
\newblock Video understanding as machine translation.
\newblock {\em arXiv preprint arXiv:2006.07203}, 2020.

\bibitem{krishna2017dense}
Ranjay Krishna, Kenji Hata, Frederic Ren, Li Fei-Fei, and Juan Carlos~Niebles.
\newblock Dense-captioning events in videos.
\newblock In {\em ICCV}, 2017.

\bibitem{krishna2017visual}
Ranjay Krishna, Yuke Zhu, Oliver Groth, Justin Johnson, Kenji Hata, Joshua
  Kravitz, Stephanie Chen, Yannis Kalantidis, Li-Jia Li, David~A Shamma, et~al.
\newblock Visual genome: Connecting language and vision using crowdsourced
  dense image annotations.
\newblock {\em International Journal of Computer Vision}, 123(1):32--73, 2017.

\bibitem{lee2018stacked}
Kuang-Huei Lee, Xi Chen, Gang Hua, Houdong Hu, and Xiaodong He.
\newblock Stacked cross attention for image-text matching, 2018.

\bibitem{lei2021less}
Jie Lei, Linjie Li, Luowei Zhou, Zhe Gan, Tamara~L Berg, Mohit Bansal, and
  Jingjing Liu.
\newblock Less is more: Clipbert for video-and-language learning via sparse
  sampling.
\newblock {\em arXiv preprint arXiv:2102.06183}, 2021.

\bibitem{lei2018tvqa}
Jie Lei, Licheng Yu, Mohit Bansal, and Tamara~L Berg.
\newblock Tvqa: Localized, compositional video question answering.
\newblock {\em arXiv preprint arXiv:1809.01696}, 2018.

\bibitem{li2020hero}
Linjie Li, Yen-Chun Chen, Yu Cheng, Zhe Gan, Licheng Yu, and Jingjing Liu.
\newblock Hero: Hierarchical encoder for video+ language omni-representation
  pre-training.
\newblock {\em EMNLP}, 2020.

\bibitem{li2020oscar}
Xiujun Li, Xi Yin, Chunyuan Li, Pengchuan Zhang, Xiaowei Hu, Lei Zhang, Lijuan
  Wang, Houdong Hu, Li Dong, Furu Wei, et~al.
\newblock Oscar: Object-semantics aligned pre-training for vision-language
  tasks.
\newblock In {\em ECCV}, 2020.

\bibitem{lin2014microsoft}
Tsung-Yi Lin, Michael Maire, Serge Belongie, James Hays, Pietro Perona, Deva
  Ramanan, Piotr Doll{\'a}r, and C~Lawrence Zitnick.
\newblock Microsoft {COCO}: Common objects in context.
\newblock In {\em ECCV}, 2014.

\bibitem{liu2021hit}
Song Liu, Haoqi Fan, Shengsheng Qian, Yiru Chen, Wenkui Ding, and Zhongyuan
  Wang.
\newblock Hit: Hierarchical transformer with momentum contrast for video-text
  retrieval, 2021.

\bibitem{Liu19a}
Yang Liu, Samuel Albanie, Arsha Nagrani, and Andrew Zisserman.
\newblock Use what you have: Video retrieval using representations from
  collaborative experts.
\newblock In {\em Proc. BMVC}, 2019.

\bibitem{gst}
Chenxu Luo and Alan Yuille.
\newblock Grouped spatial-temporal aggretation for efficient action
  recognition.
\newblock In {\em ICCV}, 2019.

\bibitem{luo2020univilm}
Huaishao Luo, Lei Ji, Botian Shi, Haoyang Huang, Nan Duan, Tianrui Li, Xilin
  Chen, and Ming Zhou.
\newblock {UniVL}: A unified video and language pre-training model for
  multimodal understanding and generation.
\newblock {\em arXiv preprint arXiv:2002.06353}, 2020.

\bibitem{mccarthy2010mtld}
Philip~M McCarthy and Scott Jarvis.
\newblock Mtld, vocd-d, and hd-d: A validation study of sophisticated
  approaches to lexical diversity assessment.
\newblock {\em Behavior research methods}, 42(2):381--392, 2010.

\bibitem{miech20endtoend}
Antoine Miech, Jean-Baptiste Alayrac, Lucas Smaira, Ivan Laptev, Josef Sivic,
  and Andrew Zisserman.
\newblock End-to-end learning of visual representations from uncurated
  instructional videos.
\newblock In {\em CVPR}, 2020.

\bibitem{miech18learning}
Antoine Miech, Ivan Laptev, and Josef Sivic.
\newblock Learning a text-video embedding from incomplete and heterogeneous
  data.
\newblock {\em arXiv}, 2018.

\bibitem{miech2019howto100m}
Antoine Miech, Dimitri Zhukov, Jean-Baptiste Alayrac, Makarand Tapaswi, Ivan
  Laptev, and Josef Sivic.
\newblock Howto100m: Learning a text-video embedding by watching hundred
  million narrated video clips.
\newblock In {\em ICCV}, 2019.

\bibitem{mithun2018learning}
Niluthpol~Chowdhury Mithun, Juncheng Li, Florian Metze, and Amit~K
  Roy-Chowdhury.
\newblock Learning joint embedding with multimodal cues for cross-modal
  video-text retrieval.
\newblock In {\em Proceedings of the 2018 ACM on International Conference on
  Multimedia Retrieval}, pages 19--27, 2018.

\bibitem{parmar2018image}
Niki Parmar, Ashish Vaswani, Jakob Uszkoreit, Lukasz Kaiser, Noam Shazeer,
  Alexander Ku, and Dustin Tran.
\newblock Image transformer.
\newblock In {\em ICML}, 2018.

\bibitem{NEURIPS2019_9015}
Adam Paszke, Sam Gross, Francisco Massa, Adam Lerer, James Bradbury, Gregory
  Chanan, Trevor Killeen, Zeming Lin, Natalia Gimelshein, Luca Antiga, Alban
  Desmaison, Andreas Kopf, Edward Yang, Zachary DeVito, Martin Raison, Alykhan
  Tejani, Sasank Chilamkurthy, Benoit Steiner, Lu Fang, Junjie Bai, and Soumith
  Chintala.
\newblock Pytorch: An imperative style, high-performance deep learning library.
\newblock In {\em NeurIPS}, 2019.

\bibitem{patrick2020support}
Mandela Patrick, Po-Yao Huang, Yuki Asano, Florian Metze, Alexander Hauptmann,
  Jo{\~a}o Henriques, and Andrea Vedaldi.
\newblock Support-set bottlenecks for video-text representation learning.
\newblock {\em arXiv preprint arXiv:2010.02824}, 2020.

\bibitem{radford2021learning}
Alec Radford, Jong~Wook Kim, Chris Hallacy, Aditya Ramesh, Gabriel Goh,
  Sandhini Agarwal, Girish Sastry, Amanda Askell, Pamela Mishkin, Jack Clark,
  Gretchen Krueger, and Ilya Sutskever.
\newblock Learning transferable visual models from natural language
  supervision, 2021.

\bibitem{ramachandran2019stand}
Prajit Ramachandran, Niki Parmar, Ashish Vaswani, Irwan Bello, Anselm Levskaya,
  and Jonathon Shlens.
\newblock Stand-alone self-attention in vision models.
\newblock {\em arXiv preprint arXiv:1906.05909}, 2019.

\bibitem{regneri2013grounding}
Michaela Regneri, Marcus Rohrbach, Dominikus Wetzel, Stefan Thater, Bernt
  Schiele, and Manfred Pinkal.
\newblock Grounding action descriptions in videos.
\newblock {\em Transactions of the Association for Computational Linguistics},
  1:25--36, 2013.

\bibitem{Rohrbach_2015_CVPR}
Anna Rohrbach, Marcus Rohrbach, Niket Tandon, and Bernt Schiele.
\newblock A dataset for movie description.
\newblock In {\em CVPR}, 2015.

\bibitem{rohrbach2017movie}
Anna Rohrbach, Atousa Torabi, Marcus Rohrbach, Niket Tandon, Christopher Pal,
  Hugo Larochelle, Aaron Courville, and Bernt Schiele.
\newblock Movie description.
\newblock {\em International Journal of Computer Vision}, 123(1):94--120, 2017.

\bibitem{rohrbach2012database}
Marcus Rohrbach, Sikandar Amin, Mykhaylo Andriluka, and Bernt Schiele.
\newblock A database for fine grained activity detection of cooking activities.
\newblock In {\em CVPR}, 2012.

\bibitem{rouditchenko2020avlnet}
Andrew Rouditchenko, Angie Boggust, David Harwath, Dhiraj Joshi, Samuel Thomas,
  Kartik Audhkhasi, Rogerio Feris, Brian Kingsbury, Michael Picheny, Antonio
  Torralba, et~al.
\newblock {AVLnet}: Learning audio-visual language representations from
  instructional videos.
\newblock {\em arXiv preprint arXiv:2006.09199}, 2020.

\bibitem{distilbert}
Victor Sanh, Lysandre Debut, Julien Chaumond, and Thomas Wolf.
\newblock Distilbert, a distilled version of {BERT:} smaller, faster, cheaper
  and lighter.
\newblock {\em CoRR}, abs/1910.01108, 2019.

\bibitem{seo2021look}
Paul~Hongsuck Seo, Arsha Nagrani, and Cordelia Schmid.
\newblock Look before you speak: Visually contextualized utterances.
\newblock {\em CVPR}, 2021.

\bibitem{sevilla2021only}
Laura Sevilla-Lara, Shengxin Zha, Zhicheng Yan, Vedanuj Goswami, Matt Feiszli,
  and Lorenzo Torresani.
\newblock Only time can tell: Discovering temporal data for temporal modeling.
\newblock In {\em WACV}, 2021.

\bibitem{sharma2018conceptual}
Piyush Sharma, Nan Ding, Sebastian Goodman, and Radu Soricut.
\newblock Conceptual captions: A cleaned, hypernymed, image alt-text dataset
  for automatic image captioning.
\newblock In {\em Proceedings of the 56th Annual Meeting of the Association for
  Computational Linguistics (Volume 1: Long Papers)}, pages 2556--2565, 2018.

\bibitem{sigurdsson2016hollywood}
Gunnar~A Sigurdsson, G{\"u}l Varol, Xiaolong Wang, Ali Farhadi, Ivan Laptev,
  and Abhinav Gupta.
\newblock Hollywood in homes: Crowdsourcing data collection for activity
  understanding.
\newblock In {\em ECCV}, 2016.

\bibitem{touvron2020deit}
Hugo Touvron, Matthieu Cord, Matthijs Douze, Francisco Massa, Alexandre
  Sablayrolles, and Herv\'e J\'egou.
\newblock Training data-efficient image transformers \& distillation through
  attention.
\newblock {\em arXiv preprint arXiv:2012.12877}, 2020.

\bibitem{Tran2018ACL}
Du Tran, Heng Wang, L. Torresani, Jamie Ray, Y. LeCun, and Manohar Paluri.
\newblock A closer look at spatiotemporal convolutions for action recognition.
\newblock In {\em CVPR}, 2018.

\bibitem{vaswani2017attention}
Ashish Vaswani, Noam Shazeer, Niki Parmar, Jakob Uszkoreit, Llion Jones,
  Aidan~N Gomez, Lukasz Kaiser, and Illia Polosukhin.
\newblock Attention is all you need.
\newblock In {\em NeurIPS}, 2017.

\bibitem{venugopalan2014translating}
Subhashini Venugopalan, Huijuan Xu, Jeff Donahue, Marcus Rohrbach, Raymond
  Mooney, and Kate Saenko.
\newblock Translating videos to natural language using deep recurrent neural
  networks.
\newblock {\em arXiv preprint arXiv:1412.4729}, 2014.

\bibitem{vinyals2016show}
Oriol Vinyals, Alexander Toshev, Samy Bengio, and Dumitru Erhan.
\newblock Show and tell: Lessons learned from the 2015 {MSCOCO} image
  captioning challenge.
\newblock {\em IEEE transactions on pattern analysis and machine intelligence},
  39(4):652--663, 2016.

\bibitem{wang2016learning}
Liwei Wang, Yin Li, and Svetlana Lazebnik.
\newblock Learning deep structure-preserving image-text embeddings.
\newblock In {\em CVPR}, 2016.

\bibitem{wang_tsn}
L. {Wang}, Y. {Xiong}, Z. {Wang}, Y. {Qiao}, D. {Lin}, X. {Tang}, and L. {Van
  Gool}.
\newblock Temporal segment networks for action recognition in videos.
\newblock {\em IEEE Transactions on Pattern Analysis and Machine Intelligence},
  41(11):2740--2755, 2019.

\bibitem{wang2018non}
Xiaolong Wang, Ross Girshick, Abhinav Gupta, and Kaiming He.
\newblock Non-local neural networks.
\newblock In {\em CVPR}, 2018.

\bibitem{wang2021t2vlad}
Xiaohan Wang, Linchao Zhu, and Yi Yang.
\newblock T2vlad: Global-local sequence alignment for text-video retrieval,
  2021.

\bibitem{Wu2020AMM}
Chao-Yuan Wu, Ross~B. Girshick, Kaiming He, Christoph Feichtenhofer, and
  Philipp Krahenbuhl.
\newblock A multigrid method for efficiently training video models.
\newblock In {\em CVPR}, 2020.

\bibitem{xie2018rethinking}
Saining Xie, Chen Sun, Jonathan Huang, Zhuowen Tu, and Kevin Murphy.
\newblock Rethinking spatiotemporal feature learning: Speed-accuracy trade-offs
  in video classification.
\newblock In {\em ECCV}, 2018.

\bibitem{xu2016msr}
Jun Xu, Tao Mei, Ting Yao, and Yong Rui.
\newblock Msr-vtt: A large video description dataset for bridging video and
  language.
\newblock In {\em CVPR}, 2016.

\bibitem{you2016image}
Quanzeng You, Hailin Jin, Zhaowen Wang, Chen Fang, and Jiebo Luo.
\newblock Image captioning with semantic attention.
\newblock In {\em CVPR}, 2016.

\bibitem{young-etal-2014-image}
Peter Young, Alice Lai, Micah Hodosh, and Julia Hockenmaier.
\newblock From image descriptions to visual denotations: New similarity metrics
  for semantic inference over event descriptions.
\newblock {\em Transactions of the Association for Computational Linguistics},
  2:67--78, 2014.

\bibitem{yu2018joint}
Youngjae Yu, Jongseok Kim, and Gunhee Kim.
\newblock A joint sequence fusion model for video question answering and
  retrieval.
\newblock In {\em ECCV}, 2018.

\bibitem{Zhai2019ClassificationIA}
Andrew Zhai and Hao-Yu Wu.
\newblock Classification is a strong baseline for deep metric learning.
\newblock In {\em BMVC}, 2019.

\bibitem{zhang2018cross}
Bowen Zhang, Hexiang Hu, and Fei Sha.
\newblock Cross-modal and hierarchical modeling of video and text.
\newblock In {\em ECCV}, 2018.

\bibitem{zhou2018towards}
Luowei Zhou, Chenliang Xu, and Jason Corso.
\newblock Towards automatic learning of procedures from web instructional
  videos.
\newblock In {\em AAAI}, 2018.

\bibitem{zhou2018end}
Luowei Zhou, Yingbo Zhou, Jason~J Corso, Richard Socher, and Caiming Xiong.
\newblock End-to-end dense video captioning with masked transformer.
\newblock In {\em CVPR}, 2018.

\bibitem{zhu2020actbert}
Linchao Zhu and Yi Yang.
\newblock Actbert: Learning global-local video-text representations.
\newblock In {\em CVPR}, 2020.

\end{thebibliography}
\end{document}